\documentclass[final]{cvpr}

\usepackage{times}
\usepackage{epsfig}
\usepackage{graphicx}
\usepackage{amsmath}
\usepackage{amssymb}
\usepackage{amsthm}
\usepackage{subfig}
\usepackage{tabularx}
\usepackage{booktabs}
\usepackage{multirow}
\usepackage{multicol}
\usepackage{enumitem}
\usepackage{array}
\usepackage{color, colortbl}
\usepackage{fixltx2e}
\usepackage{makecell}
\usepackage{threeparttable}
\usepackage{microtype}
\usepackage{lipsum}
\usepackage{pifont}
\usepackage{graphicx}

\usepackage[pagebackref=true,breaklinks=true,letterpaper=true,colorlinks,bookmarks=false]{hyperref}

\newcommand{\cZ}{\mathcal{Z}}
\newcommand{\cW}{\mathcal{W}}
\newcommand{\norm}[1]{\left\lVert#1\right\rVert}
\newcommand{\IR}{\mathbb{R}}

\newcommand{\EL}{\mathcal{L}}

\def\cvprPaperID{10601}
\def\confYear{CVPR 2021}

\begin{document}

\title{Lifting 2D StyleGAN for 3D-Aware Face Generation}

\author{Yichun Shi \quad\quad Divyansh Aggarwal  \quad\quad  Anil K. Jain\\[10pt]
Michigan State University\\[5pt]
{\tt\small \{shiyichu,aggarw49\}@msu.edu, jain@cse.msu.edu}
}


\twocolumn[{%
\captionsetup{font=small}
\renewcommand\twocolumn[1][]{#1}%
\renewcommand{\arraystretch}{0.2}
\newcommand{\mmc}[1]{\multicolumn{1}{c}{#1}}
\maketitle
\thispagestyle{empty}
\begin{center}
\vspace{-1.0em}
\small
    \centering
    \setlength{\tabcolsep}{0pt}
    \begin{tabularx}{1.0\linewidth}{c X c X ccccc X ccc}
    \includegraphics[width=0.096\linewidth]{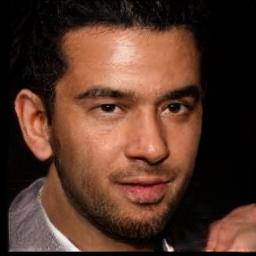} &&
    \includegraphics[width=0.096\linewidth]{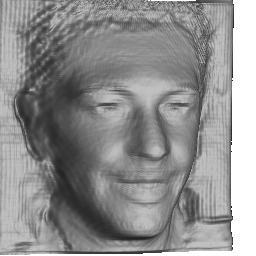} &&
    \includegraphics[width=0.096\linewidth]{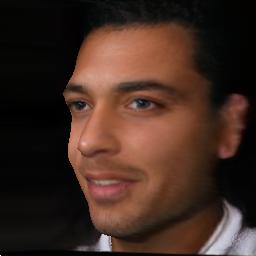} &
    \includegraphics[width=0.096\linewidth]{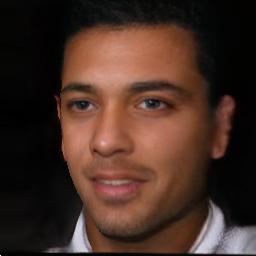} &
    \includegraphics[width=0.096\linewidth]{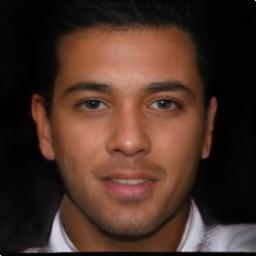} &
    \includegraphics[width=0.096\linewidth]{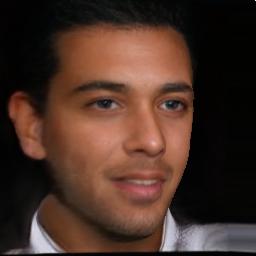} &
    \includegraphics[width=0.096\linewidth]{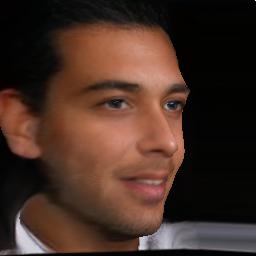} &\;&
    \includegraphics[width=0.096\linewidth]{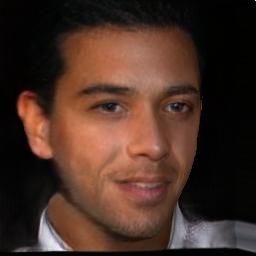} &
    \includegraphics[width=0.096\linewidth]{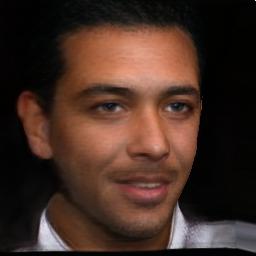} &
    \includegraphics[width=0.096\linewidth]{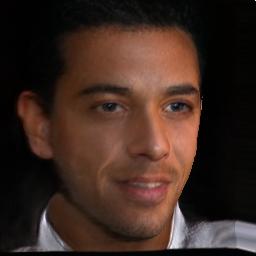} \\
    \includegraphics[width=0.096\linewidth]{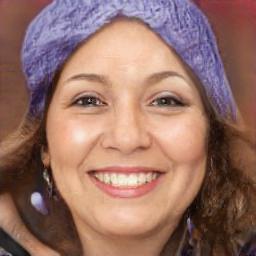} &&
    \includegraphics[width=0.096\linewidth]{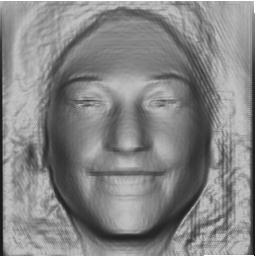} &&
    \includegraphics[width=0.096\linewidth]{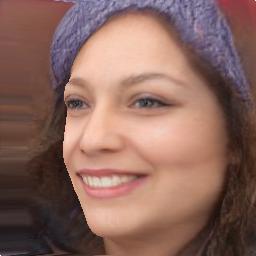} &
    \includegraphics[width=0.096\linewidth]{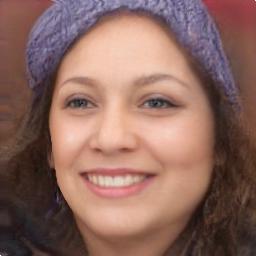} &
    \includegraphics[width=0.096\linewidth]{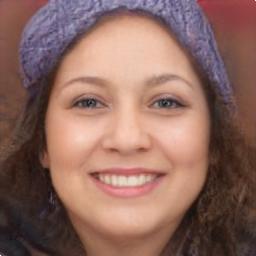} &
    \includegraphics[width=0.096\linewidth]{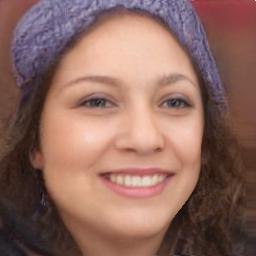} &
    \includegraphics[width=0.096\linewidth]{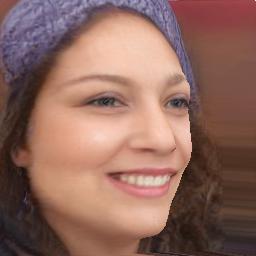} &\;&
    \includegraphics[width=0.096\linewidth]{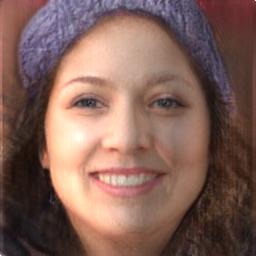} &
    \includegraphics[width=0.096\linewidth]{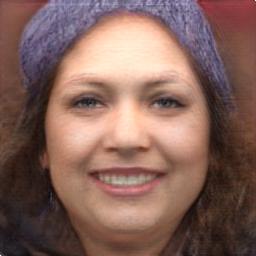} &
    \includegraphics[width=0.096\linewidth]{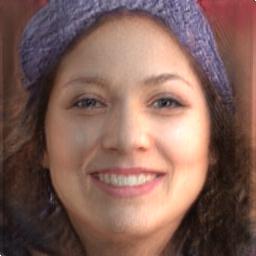} \\
    \includegraphics[width=0.096\linewidth]{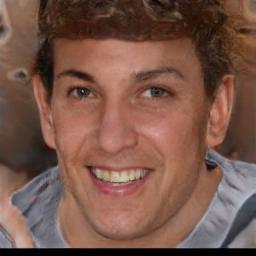} &&
    \includegraphics[width=0.096\linewidth]{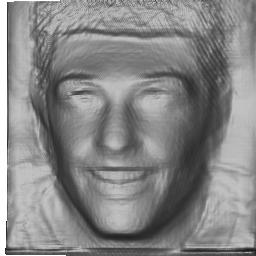} &&
    \includegraphics[width=0.096\linewidth]{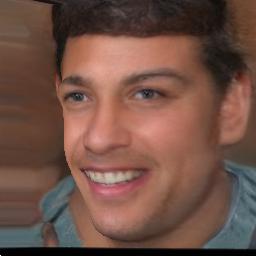} &
    \includegraphics[width=0.096\linewidth]{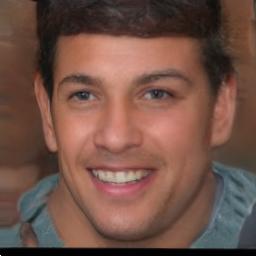} &
    \includegraphics[width=0.096\linewidth]{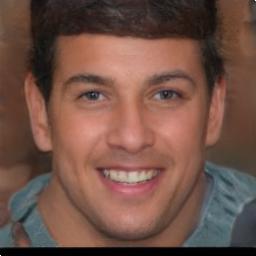} &
    \includegraphics[width=0.096\linewidth]{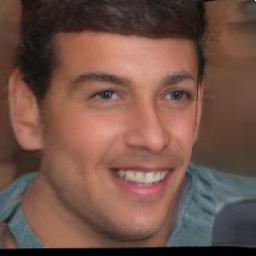} &
    \includegraphics[width=0.096\linewidth]{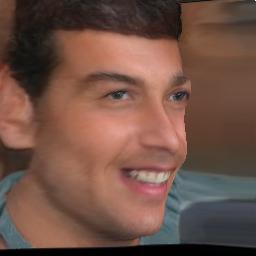} &\;&
    \includegraphics[width=0.096\linewidth]{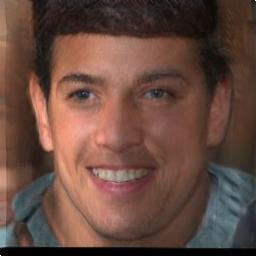} &
    \includegraphics[width=0.096\linewidth]{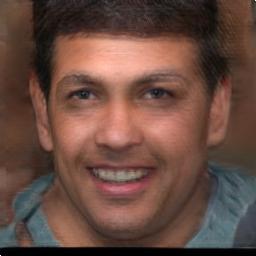} &
    \includegraphics[width=0.096\linewidth]{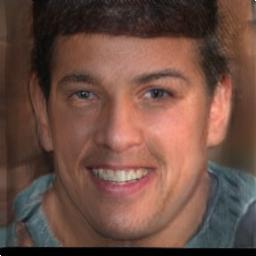} \\[0.2em]
    (a) StyleGAN2 && (b) Shape && \multicolumn{5}{c}{(c) Rotation} && \multicolumn{3}{c}{(d) Relighting} \\[-0.5em]
    \end{tabularx}
    \captionof{figure}{Example results of LiftedGAN. Column (a) shows three random samples from the latent space of a pre-trained StyleGAN2 network. Columns (b)-(d) show the results of LiftedGAN. The proposed method lifts a pre-trained StyleGAN2 to a 3D generator by predicting additional depth information, which allows it to not only generate realistic face images, but also provides 3D control of the output, such as rotation and relighting. Our method does \textbf{NOT} need any annotation nor 3DMM model for training.}
    \label{fig:frontpage}
\end{center}
}]

\begin{abstract}
We propose a framework, called LiftedGAN, that disentangles and lifts a pre-trained StyleGAN2 for 3D-aware face generation. Our model is ``3D-aware'' in the sense that it is able to (1) disentangle the latent space of StyleGAN2 into texture, shape, viewpoint, lighting and (2) generate 3D components for rendering synthetic images. Unlike most previous methods, our method is completely self-supervised, i.e. it neither requires any manual annotation nor 3DMM model for training. Instead, it learns to generate images as well as their 3D components by distilling the prior knowledge in StyleGAN2 with a differentiable renderer. The proposed model is able to output both the 3D shape and texture, allowing explicit pose and lighting control over generated images. Qualitative and quantitative results show the superiority of our approach over existing methods on 3D-controllable GANs in content controllability while generating realistic high quality images.
\end{abstract}

\section{Introduction}

Generative Adversarial Networks (GANs), such as StyleGAN~\cite{stylegan, stylegan2} have been demonstrated to generate high quality face images with a wide variety of styles. However, since these models are trained to generate random faces, they do not offer direct manipulation over the semantic attributes such as pose, expression etc. in the generated image. A number of studies have been devoted to achieving control over the generation process in order to be able to adjust pose and other semantic attributes in the generated face images. Among all these attributes, 3D information, such as pose, is the most desirable due to its applicability in face recognition~\cite{tran2017disentangled} and face synthesis~\cite{zhao2017dual}. To achieve this, most existing approaches attempt to disentangle the latent feature space of GANs by leveraging external supervision on the semantic factors such as pose labels~\cite{tran2017disentangled, tian2018cr}, landmarks~\cite{hu2018pose} or synthetic images~\cite{deng2020disentangled, zhao20183d,kowalski2020config}, while some others~\cite{nguyen2019hologan} have explored an unsupervised approach for 3D controllability in the latent space. Although these feature manipulation based methods have shown ability to generate faces with high visual quality under assigned poses, it is unclear whether important content, such as identity, is indeed preserved when we change the pose parameters. Potential errors could arise from the generation process when the manipulated features are parsed by the network parameters (see Section~\ref{sec:comparison_related}).

In contrast to solutions that only output 2D images, building generative models with explicit 3D shapes could give a stricter control of the content. For general 3D objects, a thread of recent studies train 3D generative models from 2D images, but they mostly work on rendered images with coarse shapes~\cite{henderson2019learning,henderson2020leveraging,lunz2020inverse}, e.g. cars. For faces, which contain many fine-grained details, existing solutions for 3D image generation~\cite{sahasrabudhe2019lifting,szabo2019unsupervised} have suffered from collapsed results under large pose variations due to the difficulty of learning reasonable shapes.



In this paper, we propose a framework that shares the advantages of both 2D and 3D solutions. Given a pre-trained StyleGAN2, we distill it into a 3D-aware generator, which not only outputs the generated image, but its view points, light direction and 3D information, such as surface normal map. Compared with 3D generative models, our approach is able to output rendered images with higher quality, close to 2D generative models. Compared with 2D solutions based on feature manipulation, our model allows a stricter 3D control over the content by maneuvering the view point and shading of textured meshes, as in 3D generators. Qualitative and quantitative results show the superiority of our approach over existing methods in preserving identity as well as generating realistic high quality images.


The main contributions of the paper can be summarized as follows:
\begin{itemize}[leftmargin=15pt]\vspace{-0.5em}
    \item A framework for 3D-aware face image generation, called LiftedGAN, which distills the knowledge from a pretrained StyleGAN2. \vspace{-0.5em}
    \item A self-supervised method for disentangling and distilling the 3D information in the latent space of StyleGAN2.\vspace{-0.5em}
    \item A generator that outputs both high quality face images and their 3D information, allowing explicit control over pose and lighting.\vspace{-0.5em}
\end{itemize}

\section{Related Work}
\subsection{Pose-Disentangled 2D GANs}
Recent progress in Generative Adversarial Networks  has enabled generation of high-quality realistic images. This has resulted in a body of work to disentangle different factors of the generated images from GANs. These publications can be categorized into two types. The first type explicitly adds additional modules or loss functions during training to ensure the disentanglement of pose information. For example, Tran~\etal~\cite{tran2017disentangled} and Tian~\etal~\cite{tian2018cr} use pose labels while Hu~\etal~\cite{hu2018pose} and Zhao~\etal~\cite{zhao20183d} use landmarks and a 3DMM model, respectively to guide the training of an image-translation GAN for rotating input faces. For the generation task, Deng~\etal~\cite{deng2020disentangled} use a 3DMM model during GAN training to guide the learning of a disentangled pose factor. CONFIG~\cite{kowalski2020config} mixes real face images and synthesized images from a graphic pipeline with known parameters for training the GAN. HoloGAN~\cite{nguyen2019hologan}, on the other hand, proposes to use a 3D feature projection module in the early stage of the generator to enable the rotation of the output face images. The second type, on the other hand, tries to manipulate a pre-trained GAN network to change the 3D information of the output. This is built upon the foundation that recent GANs~\cite{stylegan,stylegan2} provide a naturally disentangled latent space for generation.
Similar to training-based methods, these studies use labels~\cite{shen2020interpreting} or a 3DMM~\cite{tewari2020stylerig} to achieve the disentanglement of the pre-trained latent space. Shen~\etal~\cite{shen2020closed} have also proposed an unsupervised method to factorize the latent space of GANs. A clear drawback of all these methods is that they could not explicitly output the 3D shape of the object in the generated images, which is essential for strict 3D control over the content.

\newcommand{\tick}{\ding{52}} 

\begin{table}[!t]
\scriptsize
\setlength{\tabcolsep}{2.2pt}
\captionsetup{font=footnotesize}
\centering
\begin{threeparttable}
\renewcommand{\arraystretch}{1.3}
\begin{tabularx}{\linewidth}{X c c c c l}
\noalign{\hrule height 1.0pt}
\textbf{Study} & \textbf{Generative} & \textbf{Shape} & \textbf{Lighting} & \textbf{Supervision}\\
\noalign{\hrule height 1.0pt}




 
Tulsiani \etal~\cite{tulsiani2018multi} &  & \checkmark &  & Multi-view images\\

Kanazawa \etal~\cite{kanazawa2018learning} &  & \checkmark &  & Keypoints, Silhouette \\

Gadelha \etal~\cite{gadelha20173d} & \checkmark & \checkmark &  & Silhouette\\

Henderson \etal~\cite{henderson2018learning,henderson2019learning, henderson2020leveraging} & \checkmark & \checkmark &  & Viewpoint, Silhouette\\

Lunz \etal~\cite{lunz2020inverse} & \checkmark & \checkmark &  & None\\

Wu \etal~\cite{wu2020unsupervised} &  & \checkmark & \checkmark & None\\

Zhang \etal~\cite{zhang2020image} &  & \checkmark & \checkmark & Viewpoint\\

Pan \etal~\cite{pan2020gan2shape} &  & \checkmark & \checkmark & None\\

\noalign{\hrule height 0.5pt}

Szabo \etal~\cite{szabo2019unsupervised} & \checkmark & \checkmark &  & None\\

CONFIG~\cite{kowalski2020config} & \checkmark &  & \checkmark & Synthetic data \\

HoloGAN~\cite{nguyen2019hologan} & \checkmark &  & \checkmark & None\\

StyleRig~\cite{tewari2020stylerig} & \checkmark &  & \checkmark & 3DMM\\

DiscoFaceGAN~\cite{deng2020disentangled} & \checkmark & & \checkmark & 3DMM \\

\noalign{\hrule height 1pt}
Ours & \checkmark & \checkmark & \checkmark & None\\
\noalign{\hrule height 1pt}

\end{tabularx}
 \end{threeparttable}
\caption{Difference between our work and related work. The first half shows relevant work on unsupervised and weakly-supervised 3D reconstruction. The second half are relevant studies on 3D-controllable face generation.}\vspace{-1.2em}
\label{tab:related_work}
\end{table} 

\subsection{Unsupervised 3D Reconstruction and Generation from 2D Images}
In contrast to the 2D-based solutions above, several studies have explored the possibility of reconstructing and generating 3D shapes from unlabeled 2D images. Here, ``unlabeled" means that neither 3D shape nor view label is available during the training of the model. Such unsupervised reconstruction methods typically use a special cue to guide the learning of the 3D shape. For example, Tulsiani~\etal~\cite{tulsiani2018multi} use the muiti-view consistency as the supervision; Kanazawa~\etal~\cite{kanazawa2018learning} use the consistency between objects under the same category to learn the reconstruction model. Wu~\etal~\cite{wu2020unsupervised} use the symmetry property of the objects to learn detailed shape and albedo from natural images. Handerson~\etal~\cite{henderson2018learning,henderson2019learning,henderson2020leveraging} exploit the shading information from the synthesized images to reconstruct 3D meshes, which is further extended to a generative model by using a VAE structure. Most generative models, on the other hand, use a GAN-structure where adversarial loss provides the signal for 2D images rendered from the generated 3D shapes. Gadelha~\etal~\cite{gadelha20173d} apply the discriminator to the silhouette of the generated voxels to train the generator. Lunz~\etal~\cite{lunz2020inverse} use a commercial renderer to guide a neural renderer to output images with shading for the discriminator. However, both methods utilize voxels, which cannot recover the fine-grained details nor the colors of the 3D surfaces. Recently, Szabo~\etal~\cite{szabo2019unsupervised} proposed to use vertex position maps as the 3D representation to directly train the GAN with textured mesh outputs. However, given the large degrees of freedom of such representation and the noisy signals from adversarial training, the output shapes of their work suffer from strong distortion. Concurrent with our work, Zhang~\etal~\cite{zhang2020image} and Pan~\etal~\cite{pan2020gan2shape} have utilized StyleGAN to generate multi-view synthetic data for 3D reconstruction tasks. Zhang~\etal~\cite{zhang2020image} conduct manual annotation on offline-generated data while Pan~\etal~\cite{pan2020gan2shape} propose to iteratively synthesize data and train the reconstruction network. Different from their work, our work builds a 3D generative model by simultaneously learning to manipulate StyleGAN2 generation and estimate 3D shapes.

\newcommand{\argmax}[1]{\underset{#1}{\operatorname{arg}\,\operatorname{max}}\;}

\begin{figure*}[t]
\captionsetup{font=footnotesize}
    \centering
    \includegraphics[width=\linewidth]{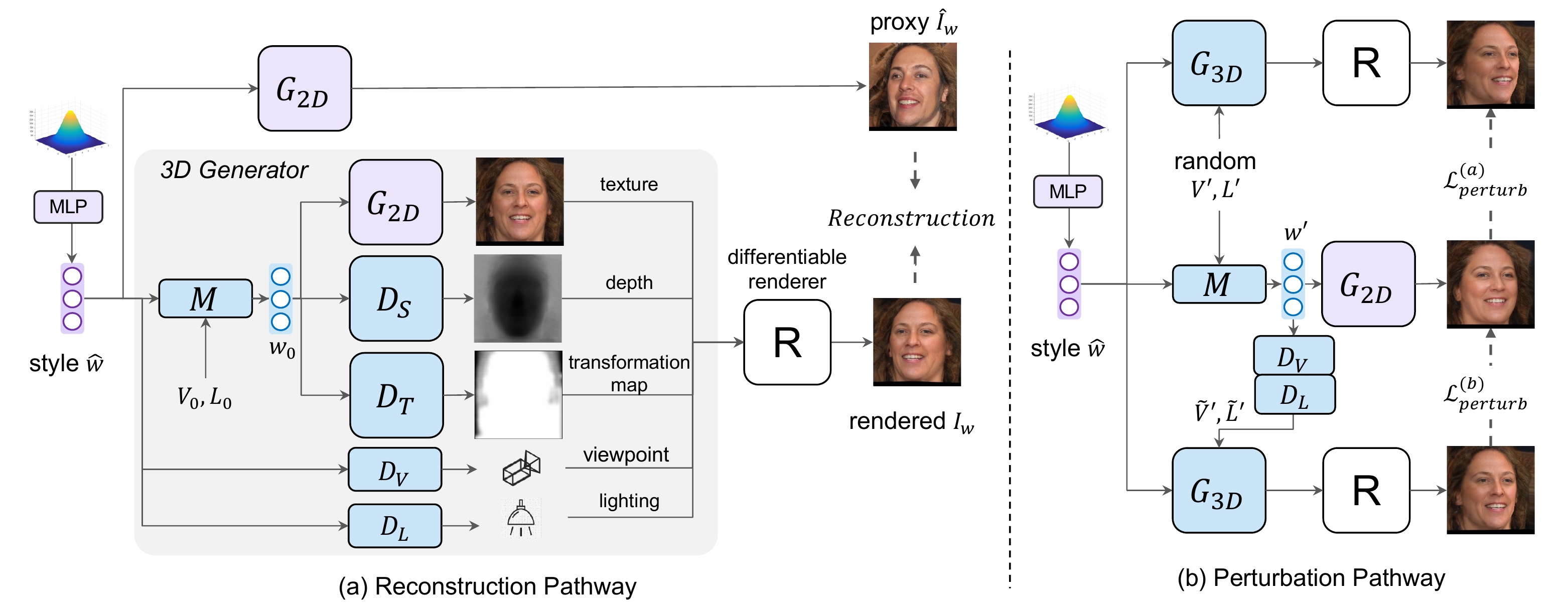}
    \caption{Overview of the proposed training framework for learning 3D generator. The framework mainly involves two pathways. First, for each randomly sampled image from StyleGAN2, we train the 3D generator to disentangle its latent code into 3D components and reconstruct the 2D image with symmetric constraint. Second, we randomly perturb the generated 3D face to obtain regularization from different views. This is achieved by simultaneously training the 3D components along with the style manipulation network, which creates pseudo ground-truth via the 2D generator to regularize the 3D face. The purple blocks in the figure indicate modules from the pre-trained StyleGAN2 and are not updated. The blue blocks are the modules to be trained. The texture images are relighted for rendering.}
    \label{fig:generator}
\end{figure*}

\section{Methodology}
The main idea of our method is to train a 3D generative network by distilling the knowledge in StyleGAN2. The StyleGAN2 network is composed of two parts, a multi-layer perceptron (MLP) that maps a latent code $z\in \cZ$ to a style code $w\in \cW$ and a 2D generator $G_{2D}$ that synthesizes a face image from the style code $w$. Our goal is to build a 3D generator that disentangles the generation process of $G_{2D}$ into different 3D modules, including texture, shape, lighting and pose, which are then utilized to render a 2D face image.

\subsection{3D Generator}
As shown in Figure~\ref{fig:generator}, the 3D generator, denoted as $G_{3D}$ is composed of five additional networks: $D_{V}$, $D_{L}$, $D_{S}$, $D_{T}$ and $M$. Given a randomly sampled style code $\hat{w}$, the network $D_{V}$ is an MLP that maps $\hat{w}$ to a 6-dimensional viewpoint representation $V$, including translation and rotation. $D_{L}$ is another MLP that decodes $\hat{w}$ into a 4-dimensional output $L$: the x-y direction of the light, ambient light and diffuse light intensity. The style manipulation network $M$ is a core module in our work. It serves to transfer a style code $\hat{w}$ to a new style code with specified light and view. In particular for 3D generator, it is used to create $w_0=M(\hat{w},L_0,V_0)$ where $L_0$ and $V_0$ are default parameters of neutralized lighting and viewpoint. Thus, StyleGAN2 $G_{2D}(w_0)$ outputs a neutralized face serving as texture map, as shown in Figure~\ref{fig:generator}. This neutralized face is then de-lighted by $L_0$ under a Lambertain model to obtain the albedo map $A$, where we divide the texture values by light intensity. $D_{S}$ and $D_{T}$ are two de-convolution networks that map the disentangled style code $w_0$ to the shape representation $S$ and a transformation map $T$, which are further explained in Section~\ref{sec:method:shape}. Finally, a differentiable renderer $R$ is used to output a rendered image $I_w=R(A,S,T,V,L)$.

\subsubsection{Shape Representation}
\label{sec:method:shape}
Prior work on 3D generation~\cite{szabo2019unsupervised,henderson2020leveraging} use 3D position maps to represent the mesh of the target object. The advantage of such an approach is that it could possibly disentangle the foreground and background. However, specifically for face, whose contour is often ambiguous given the irregular shape of hair, we found that forcing a separated foreground could easily lead to collapsed shapes on the boundary, which is also observed in~\cite{szabo2019unsupervised}. Therefore instead, we use a depth map with a transformation map to represent the shape. The depth map is associated with the texture map to represent their positions, while the transformation map decides how much each pixel should be transformed when we rotate the face. Formally, during face manipulation, for each pixel $(i,j)$, whose original position and target position is given by $p^{(old)}_{i,j}$ and $p^{(tgt)}_{i,j}$, the new target position of the pixel using the transformation map would be:
\begin{equation}
    p^{(new)}_{i,j} = (1-T_{i,j})p^{(old)}_{i,j} + T_{i,j}p^{(tgt)}_{i,j},
\end{equation}
where $T_{i,j}$ is the corresponding value of pixel $(i,j)$ on the transformation map $T$. All the foreground will be assigned with $T_{i,j}=1$, while border pixels are forced to have $0$ transformation freedom. The usage of transformation map allows us to dynamically transform the pixels to obtain a complete image without disentangling the background. See Figure~\ref{fig:generation_multiview} (d) for example effect of transformation map.



\subsection{Loss Functions}
Unlike other GAN-based methods~\cite{szabo2019unsupervised}, our method does not need any adversarial training. Since the relationship between the style space $\cW$ and the image space are fixed by $G_{2D}$, we can directly use $G_{2D}$ as a teacher model to supervise the $G_{3D}$ for learning the image mapping.  Then, we further use the symmetric constraint and multi-view data created by $G_{2D}$ to enable the learning of 3D shape.

\subsubsection{Reconstruction Loss}
\label{sec:method:reconstruction}
Let $\hat{w}$ be a randomly sampled code and $\hat{I}_w=G_{2D}(w)$ is a proxy image output by StyleGAN2. The rendered image is given by $I_w=R(A,S,T,V,L)$. The reconstruction loss for each sample is then defined as:
\begin{align}
\begin{split}
    \EL_{rec} = \norm{I_w-\hat{I}_w}_1 + \lambda_{perc} \EL_{perc}(I_w,\hat{I}_w),
\end{split}
\end{align}
where the second term refers to the perceptual loss~\cite{johnson2016perceptual} using a pre-trained VGG-16 network. Inspired by Wu et al.~\cite{wu2020unsupervised}, we also add an symmetric reconstruction loss $\EL_{flip}$ to reconstruct the proxy image $\hat{I}_w$. $\EL_{flip}$ has the same formulation as $\EL_{rec}$ except that it uses flipped albdeo and shape maps during the rendering. We found such a symmetric regularization to be very helpful to construct a frontalized face as texture map.

\begin{figure}[t]
\captionsetup{font=footnotesize}
\centering
\footnotesize
\setlength\tabcolsep{1px}
\newcommand{\www}{0.19\linewidth}
\newcolumntype{Y}{>{\centering\arraybackslash}X}
\begin{tabularx}{\linewidth}{Y>{\centering\arraybackslash}c}
    \raisebox{1.0\height}{\rotatebox[origin=c]{90}{StyleGAN2}} & 
    \includegraphics[width=\www]{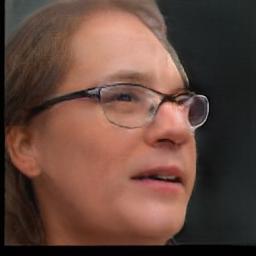}\hfill
    \includegraphics[width=\www]{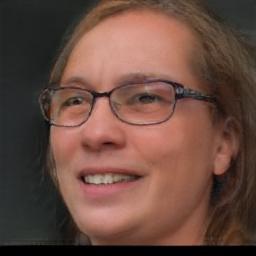}\hfill
    \includegraphics[width=\www]{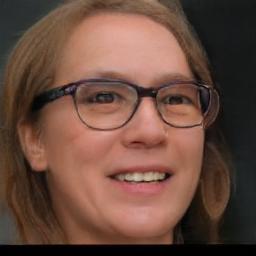}\hfill
    \includegraphics[width=\www]{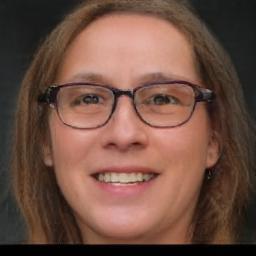}\hfill
    \includegraphics[width=\www]{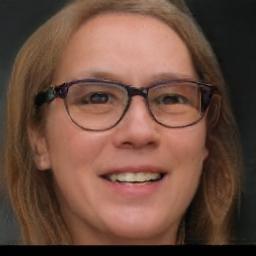}\\[-0.2em]
    \raisebox{0.8\height}{\rotatebox[origin=c]{90}{3D Generator}} & 
    \includegraphics[width=\www]{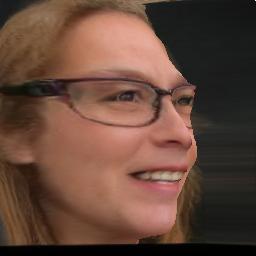}\hfill
    \includegraphics[width=\www]{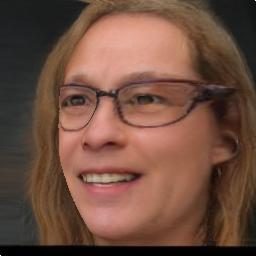}\hfill
    \includegraphics[width=\www]{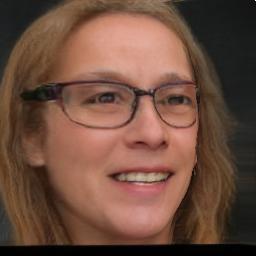}\hfill
    \includegraphics[width=\www]{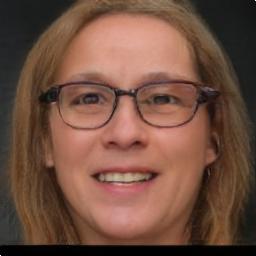}\hfill
    \includegraphics[width=\www]{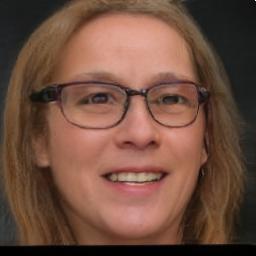}\\
\end{tabularx}
    \caption{Example images for Section~\ref{sec:multiview_generator}. The first row shows generated images of StyleGAN2 by manipulating the same style code with style manipulation network $M$. The second row shows rotated images with the differentiable render. The generated images of StyleGAN2 provides pseudo ground-truth for multi-view supervision. Note that during training, we only perturb once for each image.}
    \label{fig:generation_perturbation}
\end{figure}

\subsubsection{Generator as Multi-view Supervision}
\label{sec:multiview_generator}
The proxy images generated by $G_{2D}$ provide supervision for learning the mapping from latent space to image space. However, learning 3D shapes remains an ill-posed problem. Although we employ the symmetric reconstruction loss~\cite{wu2020unsupervised}, we found it insufficient to regularize the output shape, possibly due to the larger area of background and higher resolution of our training data. 
Here, since we already have a pre-trained generator, we utilize the $G_{2D}$ as a multi-view data generator to provide supervision for rotated-views of the 3D face. This approach is inspired by recent findings that StyleGAN2 is able to generate different views of a target sample by changing its style code~\cite{shen2020interpreting,shen2020closed,zhang2020image}. 

Formally, for each rendered sample $I_w=R(A,S,T,V,L)$, we randomly sample a different view point $V'$ and lighting $L'$ to render a rotated and relighted face image $I'_w=R(A,S,T,V',L')$. The objective here is to maximize the likelihood of this rotated face, $\log p(I'_w)$ for any random $V'$ and light $L'$ to make sure it look like a real face. Assuming the 2D generator is well trained to approximate the real data distribution, we could use the generator to estimate this likelihood. However, directly optimizing the likelihood is non-trivial, since it involves marginalizing over the latent space of $G_{2D}$. Here, given the style manipulation network $M$, for each perturbed image $I'_w$ we have a corresponding $w'=M(\hat{w},V',L')$, where we manipulate the original face in the latent space. If both the network $M$ and the shape decoder $D_S$ have well learned, $I'_w$ and $w'$ should match each other after perturbation. Thus, we optimize the joint probability $p(I'_w,w')=p_{G_{2D}}(I'_w|w')p(w')$, which is equivalent to minimizing the following loss function:
\begin{align}
\label{eq:perturb_original}
\begin{split}
    \EL'_{perturb} = = &\; - \log p_{G_{2D}}(I'_w|w') - \log p(w')\\
                = &\; d(I'_w, G_{2D}(w')) + \beta\frac{\norm{w'-\mu_w}^2}{2\sigma^2_w}
\end{split}
\end{align}\vspace{-0.5em}

Here, the prior $p(w')$ is approximated by a Gaussian distribution $\mathcal{N}(w;\mu_w,\sigma^2_w\mathbf{I})$, where empirical mean $\mu_w$ and standard deviation $\sigma_w$ of randomly generated style codes are used.
Equation~(\ref{eq:perturb_original}) can be understood as follows: for a randomly perturbed image $R(A,S,T,V',L')$,  we use StyleGAN2 to synthesize proxy images $G_{2D}(w')$ to provide pseudo ground-truth for the target view and lighting while we train the network $M$ to learn geometric warping and relighting in the latent space. The second term regularizes the transferred $w$ to be close to the prior distribution to ensure the quality of generated images. We note that $\log p(I'_w,w')$ can also be regarded as an approximation of the variational lower bound of $\log p(I'_w)$ (see supplementary). Thus, we are indirectly forcing each perturbed image to look realistic.

In practice, we found that directly optimizing Equation~(\ref{eq:perturb_original}) would flatten the output shape. Given the outputs of $G_{2D}$, we observe that the problem is caused by (1) pose difference between the generated image $G_{2D}(w)$ and the target $V'$, (2) the incapability of $G_{2D}$ to synthesize images with diverse lighting. Thus, inspired by Pan~\etal~\cite{pan2020gan2shape}, we implement Equation~(\ref{eq:perturb_original}) as two parts. The original $I'_w$ is only used for optimizing the style manipulation network $w'$ while $A$, $S$ and $T$ are optimized with re-estimated $\tilde{V}'=D_{V}(w')$ and $\tilde{L}'=D_{L}(w')$. The loss is given by:
\vspace{-0.5em}
\begin{align}
\label{eq:perturbation_separate}
\begin{split}
    \EL_{perturb} &= \EL^{(a)}_{perturb} + \EL^{(b)}_{perturb}  \\
    \EL^{(a)}_{perturb} &= d(I'_{w}, G_{2D}(z_{perturb}))  + \beta\frac{\norm{w'-\mu_w}^2}{2\sigma^2_w}\\
    \EL^{(b)}_{perturb} &= d(R(A,S,T,\tilde{V}',\tilde{L}'), \hat{I}_{w'}) + \lambda_{LV_{cyc}}\EL_{{LV}_{cyc}}, \\
\end{split}
\end{align}\vspace{-0.5em}
where 
\vspace{-0.6em}\begin{equation}\vspace{-0.6em}
\EL_{{LV}_{cyc}}=\norm{\tilde{V}'-V'}^2+\norm{\tilde{L}'-L'}^2,
\end{equation}
$I'_{w}=R(A,S,T,V',L')$ and $\hat{I}_{w'}=G_{2D}(w')$ are used as proxy images that are not updated. Our method shares the similar concept with Zhang~\etal ~\cite{zhang2020image} and Pan~\etal~\cite{pan2020gan2shape}, which use StyleGAN2 to create synthetic training data. However, different from them, we do not need any manual annotation~\cite{zhang2020image} nor iterative training~\cite{pan2020gan2shape}. All the modules are trained in an end-to-end manner. An illustration of Equation~\ref{eq:perturbation_separate} can be found in Figure~\ref{fig:generator}.  Figure~\ref{fig:generation_perturbation} shows example proxy images generated by StyleGAN2 and rendered images with re-estimated parameters.

\subsubsection{Regularization Losses}


We also add a few regularization losses to constrain the output of our model. First, assuming that the identity shouldn't change after viewpoint perturbation, we regularize the identity variance loss:
\begin{equation}
\label{eq:reg_a}
    \EL_{idt} = \norm{f(I_{w_0})-f(I'_{w})}^2,
\end{equation}
where $I_{w_0}$ is the texture map and $f$ is a pre-trained face recognition network. Since the recognition network is not guaranteed to be pose invariant, we only apply this to images that are rotated within a certain range.

Second, in order to better utilize the shading information, we regularize the albedo maps, which is acquired by delighting the canonical face image:
\begin{equation}
\label{eq:reg_a}
    \EL_{reg_A} = \norm{K_A}_{*}.
\end{equation}
Here $K_A\in \IR^{B\times HW}$ is the albedo matrix that is composed of filtered and vectorized albedo maps and $\norm{\cdot}_{*}$ denotes the nuclear norm. Laplacian kernel is used for filtering the gray-scaled albedo maps to only keep the high-frequency information. The nuclear norm is a soft approximation of low-rank regularization, which enforces different albedo maps in a batch to have smaller laplacians while encouraging consistency across samples.


The overall loss function for training the 3D generator is:
\begin{align}
    \EL_{G_{3D}} = &\;\lambda_{rec} {\EL_{rec}} + \lambda_{flip} {\EL_{flip}} + \lambda_{perturb} {\EL_{perturb}} \\ 
    &\; \lambda_{idt} {\EL_{idt}} + \lambda_{reg_A} {\EL_{reg_A}}
\end{align}

\section{Implementation Details}
We implement all the modules in this paper using Pytorch 1.5. The mesh rasterizer in Pytorch3D~\cite{ravi2020pytorch3d} is used for differentiable rendering. The StyleGAN2~\cite{stylegan2} is trained on FFHQ dataset~\cite{stylegan} with a Pytorch re-implentation~\footnote{https://github.com/rosinality/stylegan2-pytorch}. The training images are cropped and resized to 256x256 with the MTCNN face detector~\cite{MTCNN}. The 2D StyleGAN2 is then used to train the 3D generator in the second stage. The hyper-parameters $\lambda_{rec}$, $\lambda_{perc}$, $\lambda_{flip}$, $\lambda_{perturb}$, $\beta$, $\lambda_{LV_{cyc}}$, $\lambda_{idt}$ and $\lambda_{reg_A}$ are set to $5.0$, $1.0$, $0.8$, $2.0$, $0.5$, $2.0$, $1.0$ and $0.01$, respectively. These hyper-parameters are chosen based on the qualitative results of generated samples during training. Due to the space limit, more implementation details, including the network architectures are provided in the \textit{supplementary material}. 
\section{Experiments}
\subsection{Qualitative Results}
\label{sec:exp:generation}

\begin{figure}[t]
\captionsetup{font=small}
\centering
\footnotesize
\setlength\tabcolsep{1px}
\newcommand{\www}{0.19\linewidth}
\renewcommand{\arraystretch}{0.1}
\newcolumntype{Y}{>{\centering\arraybackslash}X}
\begin{tabularx}{\linewidth}{Y>{\centering\arraybackslash}c}
    \raisebox{0.9\height}{\rotatebox[origin=c]{90}{(a) StyleGAN2}} & 
    \includegraphics[width=\www]{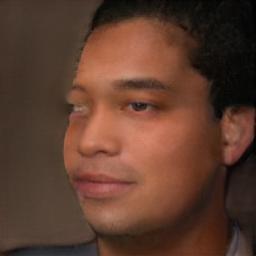}\hfill
    \includegraphics[width=\www]{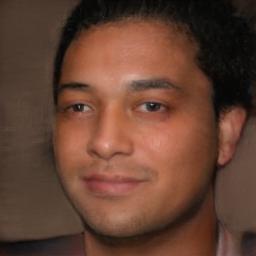}\hfill
    \includegraphics[width=\www]{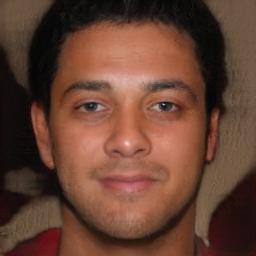}\hfill
    \includegraphics[width=\www]{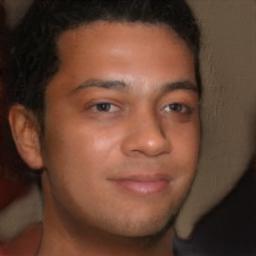}\hfill
    \includegraphics[width=\www]{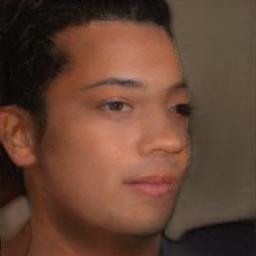}\\
    \raisebox{1.4\height}{\rotatebox[origin=c]{90}{(b) Yaw}} & 
    \includegraphics[width=\www]{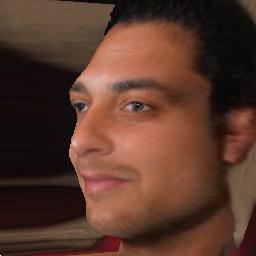}\hfill
    \includegraphics[width=\www]{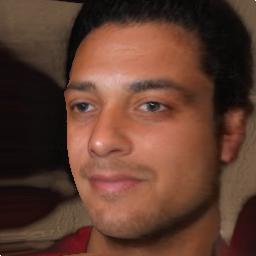}\hfill
    \includegraphics[width=\www]{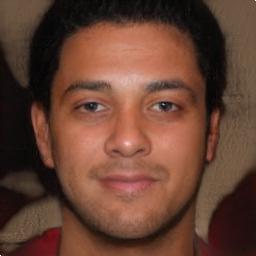}\hfill
    \includegraphics[width=\www]{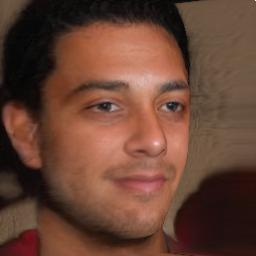}\hfill
    \includegraphics[width=\www]{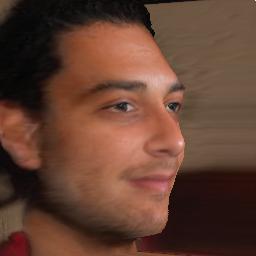}\\
    \raisebox{1.2\height}{\rotatebox[origin=c]{90}{(c) Pitch}} & 
    \includegraphics[width=\www]{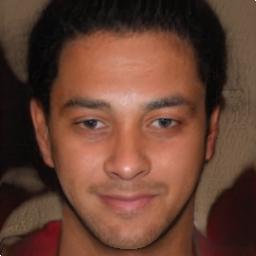}\hfill
    \includegraphics[width=\www]{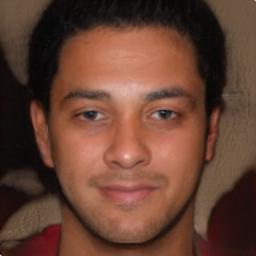}\hfill
    \includegraphics[width=\www]{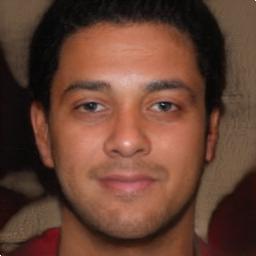}\hfill
    \includegraphics[width=\www]{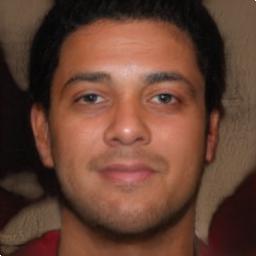}\hfill
    \includegraphics[width=\www]{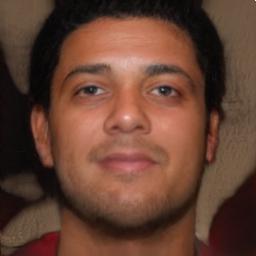}\\
    \raisebox{0.8\height}{\rotatebox[origin=c]{90}{(d) Side view}} & 
    \includegraphics[width=\www]{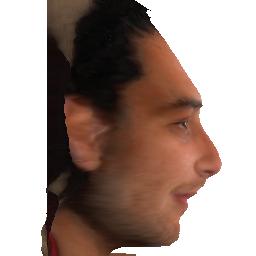}\hfill
    \includegraphics[width=\www]{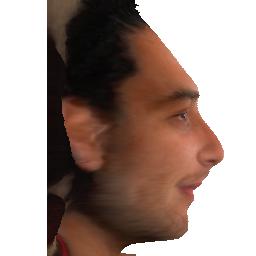}\hfill
    \includegraphics[width=\www]{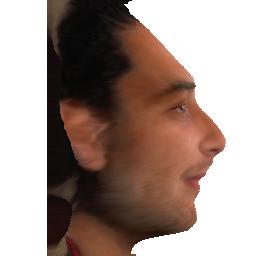}\hfill
    \includegraphics[width=\www]{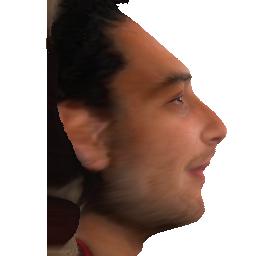}\hfill
    \includegraphics[width=\www]{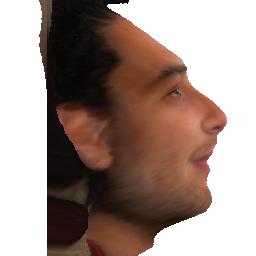}\\
\end{tabularx}
    \caption{Rotation of generated faces. Row (a) shows the results of manipulating the style code with network $M$. Row (b) and row (c) show the manipulation results of the 3D generator. Row (d) shows the rotation process from a side view.}
    \label{fig:generation_multiview}
\end{figure}

\begin{figure}[t]
\captionsetup{font=small}
\vspace{-1.0em}
    \centering
    \subfloat[Original]{\begin{minipage}{0.19\linewidth}
        \includegraphics[width=\linewidth]{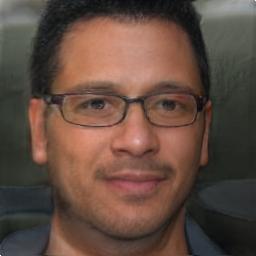}\\
        \includegraphics[width=\linewidth]{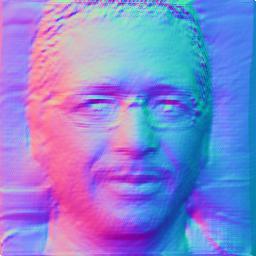}
    \end{minipage}}
    \subfloat[Left]{\begin{minipage}{0.19\linewidth}
        \includegraphics[width=\linewidth]{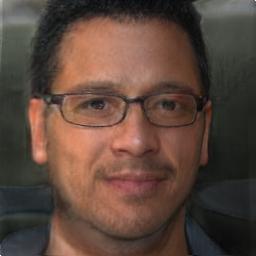}\\
        \includegraphics[width=\linewidth]{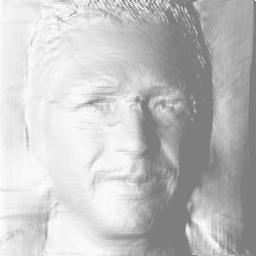}
    \end{minipage}}
    \subfloat[Top]{\begin{minipage}{0.19\linewidth}
        \includegraphics[width=\linewidth]{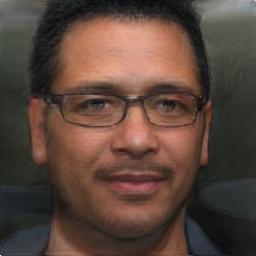}\\
        \includegraphics[width=\linewidth]{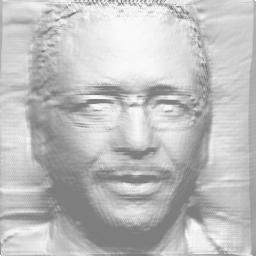}
    \end{minipage}}
    \subfloat[Right]{\begin{minipage}{0.19\linewidth}
        \includegraphics[width=\linewidth]{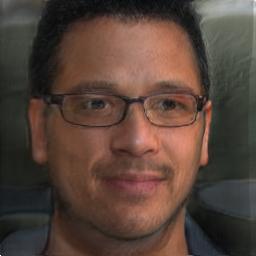}\\
        \includegraphics[width=\linewidth]{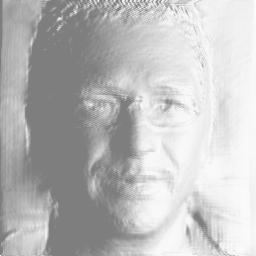}
    \end{minipage}}
    \subfloat[Bottom]{\begin{minipage}{0.19\linewidth}
        \includegraphics[width=\linewidth]{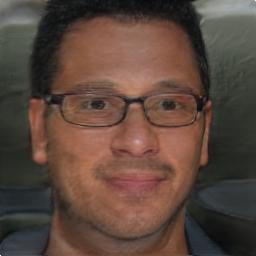}\\
        \includegraphics[width=\linewidth]{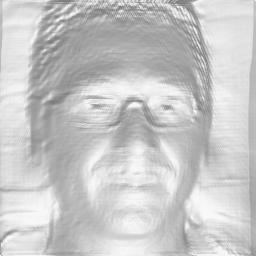}
    \end{minipage}}
    \caption{Example generated faces with different lighting. Column (a) shows a randomly sampled image and its normal map. Columns (b)-(d) show relighted images and their corresponding shading by changing the source direction of light.}
    \label{fig:generation_relight}
\end{figure}

Figure~\ref{fig:generation_multiview} shows a few examples of controlling the pose of the generated face image. The first row shows the results of manipulating the style code by network $M$, while the following rows show the results of our 3D generator. It could be seen that although latent manipulation is able to change the viewpoint of a specific face, it fails to generalize to larger poses, that are rarely seen in the training data. In contrast, the 3D generator, after distilling the 3D shape, is able to generalize to larger poses. More comparison between the style manipulation and 3D generator can be found in the supplementary material. The last row shows the side view to see how the face is rotated. With the self-predicted transformation map, only the foreground is moving while the background remains relatively static.

Figure~\ref{fig:generation_relight} shows an example generated image with different lighting conditions. By computing the normal map from the generated shape, we are able to generate face images with arbitrary lighting conditions, even those that were not seen in the original training dataset. In the second row of Figure~\ref{fig:generation_relight}, we show how the lighting from left, top, right and bottom creates the shading map, which results in the relighted images in the first row.

\begin{figure}
\captionsetup{font=small}
\centering
\footnotesize
\setlength\tabcolsep{1px}
\renewcommand{\arraystretch}{0.1}
\newcommand{\www}{0.157\linewidth}
\newcolumntype{Y}{>{\centering\arraybackslash}X}
    \begin{tabularx}{\linewidth}{Y>{\centering\arraybackslash}c}
    \raisebox{1.15\height}{\rotatebox[origin=c]{90}{Frontal}} & 
    \includegraphics[width=\www]{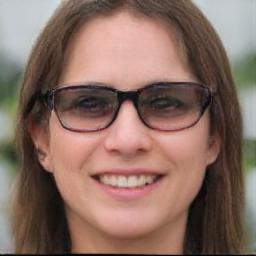}\hfill
    \includegraphics[width=\www]{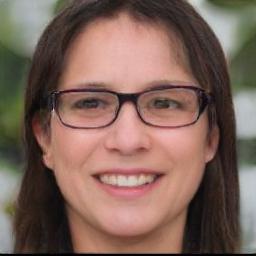}\hfill
    \includegraphics[width=\www]{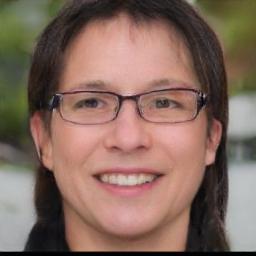}\hfill
    \includegraphics[width=\www]{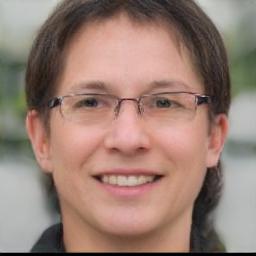}\hfill
    \includegraphics[width=\www]{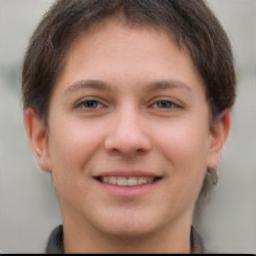}\hfill
    \includegraphics[width=\www]{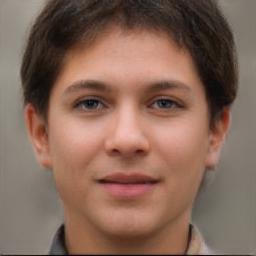}\\
    \raisebox{1.1\height}{\rotatebox[origin=c]{90}{Rotated}} & 
    \includegraphics[width=\www]{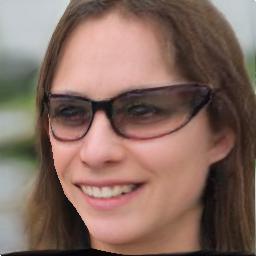}\hfill
    \includegraphics[width=\www]{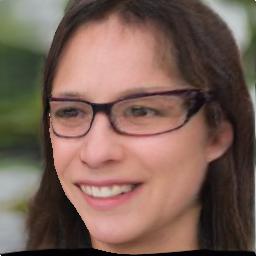}\hfill
    \includegraphics[width=\www]{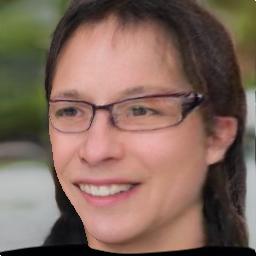}\hfill
    \includegraphics[width=\www]{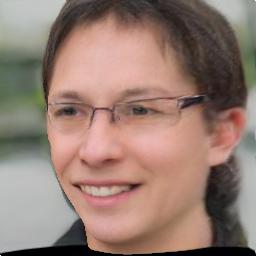}\hfill
    \includegraphics[width=\www]{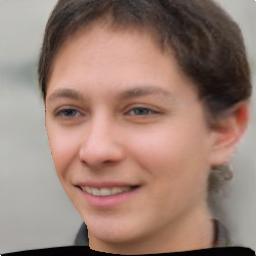}\hfill
    \includegraphics[width=\www]{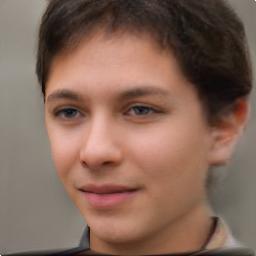}\\
    \raisebox{1.3\height}{\rotatebox[origin=c]{90}{Shape}} & 
    \includegraphics[width=\www]{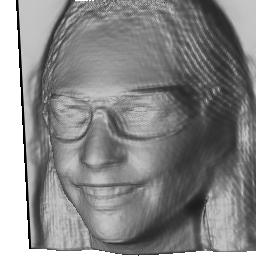}\hfill
    \includegraphics[width=\www]{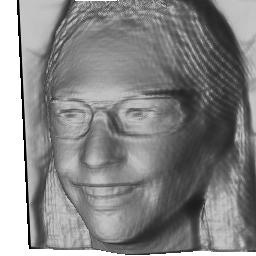}\hfill
    \includegraphics[width=\www]{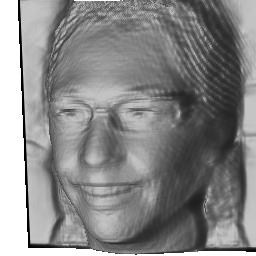}\hfill
    \includegraphics[width=\www]{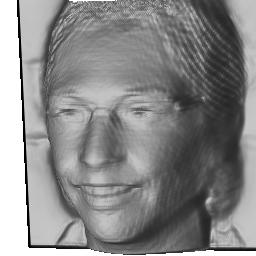}\hfill
    \includegraphics[width=\www]{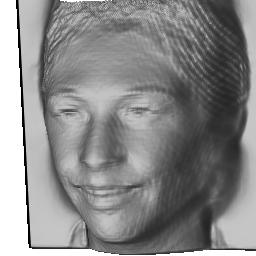}\hfill
    \includegraphics[width=\www]{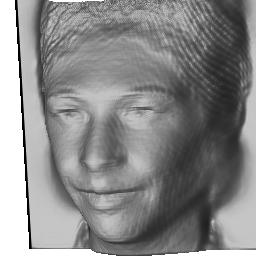}\\
    \end{tabularx}
    \vspace{-1.0em}
    \caption{Example generation results between interpolated latent codes. Our model is able to achieve a smooth change between two disparate samples, indicating its potential to generate a diverse set of controllable face images. }
    \label{fig:generation_interpolation}
\end{figure}

A successful generative model should be able to provide smooth interpolated results between disparate samples to indicate that the model is not simply memorizing training samples~\cite{stylegan}. In Figure~\ref{fig:generation_interpolation}, we show interpolated results between two faces. It could be seen that, by inheriting the style-image mapping from the 2D generator, our model is able to smoothly move from one face to another while generating realistic faces. This indicates that our model could potentially be used to generate a diverse set of faces with more effective samples compared to the original training set.

\subsection{Comparison with Related Work}
\label{sec:comparison_related}

\begin{figure}[t]
\captionsetup{font=small}
\small
\centering
\vspace{-1.0em}
\subfloat[Szabo~\etal~\cite{szabo2019unsupervised}]{\begin{minipage}{0.49\linewidth}
    \includegraphics[width=0.33\linewidth]{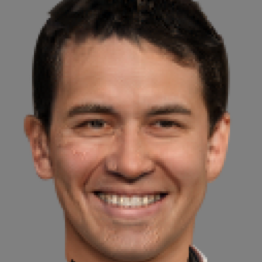}\hfill
    \includegraphics[width=0.33\linewidth]{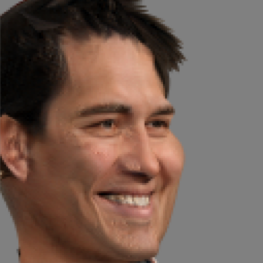}\hfill
    \includegraphics[width=0.33\linewidth]{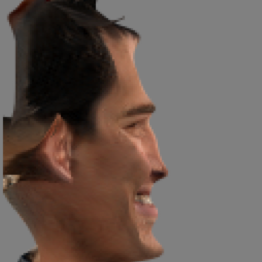}
\end{minipage}}\hfill
\subfloat[StyleRig~\cite{tewari2020stylerig}]{\begin{minipage}{0.49\linewidth}
    \includegraphics[width=0.33\linewidth]{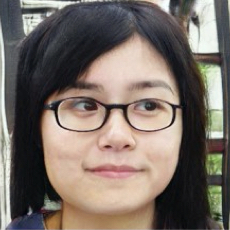}\hfill
    \includegraphics[width=0.33\linewidth]{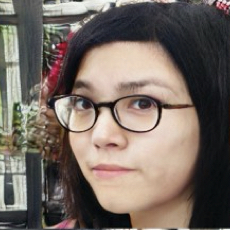}\hfill
    \includegraphics[width=0.33\linewidth]{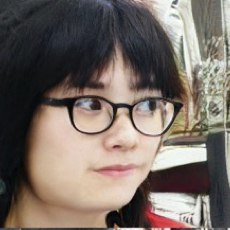}
\end{minipage}}\\[-0.01em]
\subfloat[HoloGAN~\cite{nguyen2019hologan}]{\begin{minipage}{\linewidth}
    \includegraphics[width=0.142\linewidth]{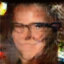}\hfill
    \includegraphics[width=0.142\linewidth]{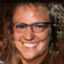}\hfill
    \includegraphics[width=0.142\linewidth]{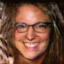}\hfill
    \includegraphics[width=0.142\linewidth]{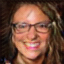}\hfill
    \includegraphics[width=0.142\linewidth]{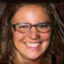}\hfill
    \includegraphics[width=0.142\linewidth]{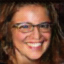}\hfill
    \includegraphics[width=0.142\linewidth]{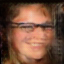}\hfill
\end{minipage}}\\[-0.01em]
\subfloat[DiscoFaceGAN~\cite{deng2020disentangled}]{\begin{minipage}{\linewidth}
    \includegraphics[width=0.142\linewidth]{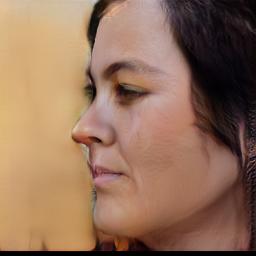}\hfill
    \includegraphics[width=0.142\linewidth]{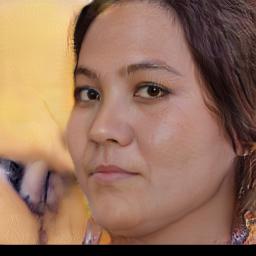}\hfill
    \includegraphics[width=0.142\linewidth]{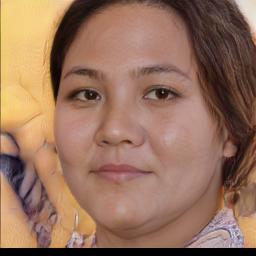}\hfill
    \includegraphics[width=0.142\linewidth]{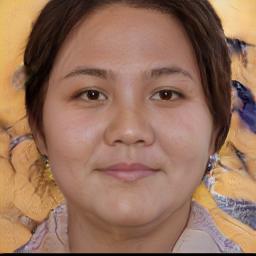}\hfill
    \includegraphics[width=0.142\linewidth]{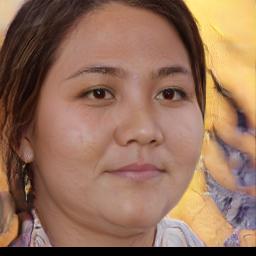}\hfill
    \includegraphics[width=0.142\linewidth]{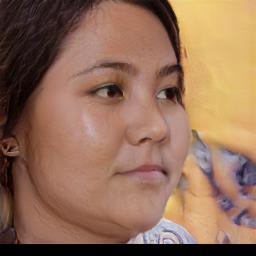}\hfill
    \includegraphics[width=0.142\linewidth]{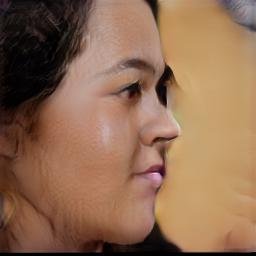}\hfill
\end{minipage}}\\[-0.01em]
\subfloat[CONFIGNet~\cite{kowalski2020config}]{\begin{minipage}{\linewidth}
    \includegraphics[width=0.142\linewidth]{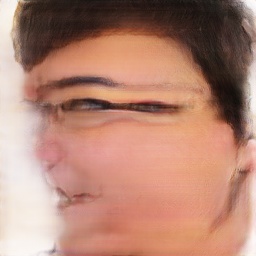}\hfill
    \includegraphics[width=0.142\linewidth]{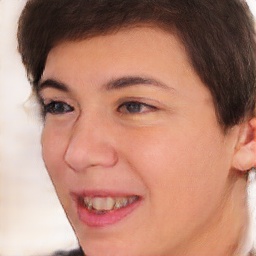}\hfill
    \includegraphics[width=0.142\linewidth]{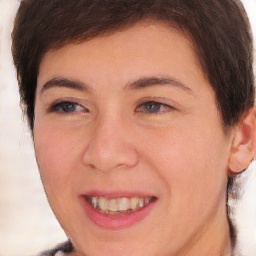}\hfill
    \includegraphics[width=0.142\linewidth]{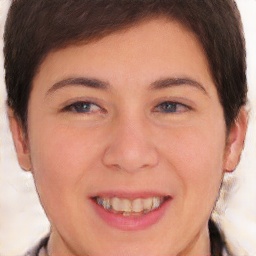}\hfill
    \includegraphics[width=0.142\linewidth]{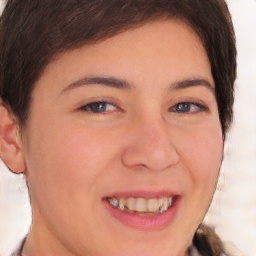}\hfill
    \includegraphics[width=0.142\linewidth]{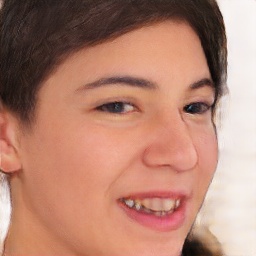}\hfill
    \includegraphics[width=0.142\linewidth]{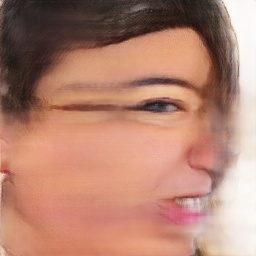}\hfill
\end{minipage}}\\[-0.01em]
\subfloat[Ours]{\begin{minipage}{\linewidth}
    \includegraphics[width=0.142\linewidth]{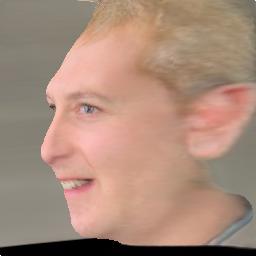}\hfill
    \includegraphics[width=0.142\linewidth]{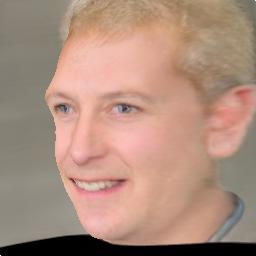}\hfill
    \includegraphics[width=0.142\linewidth]{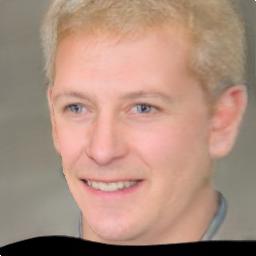}\hfill
    \includegraphics[width=0.142\linewidth]{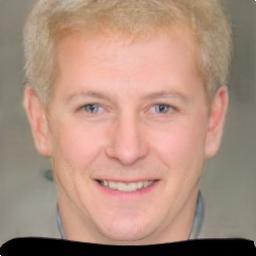}\hfill
    \includegraphics[width=0.142\linewidth]{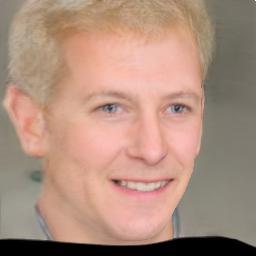}\hfill
    \includegraphics[width=0.142\linewidth]{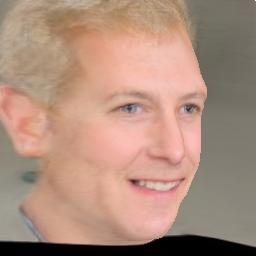}\hfill
    \includegraphics[width=0.142\linewidth]{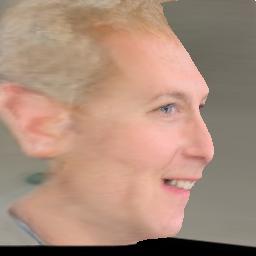}\\
    \includegraphics[width=0.142\linewidth]{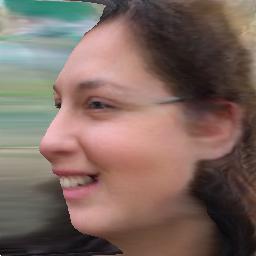}\hfill
    \includegraphics[width=0.142\linewidth]{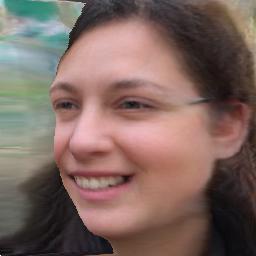}\hfill
    \includegraphics[width=0.142\linewidth]{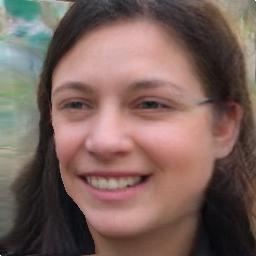}\hfill
    \includegraphics[width=0.142\linewidth]{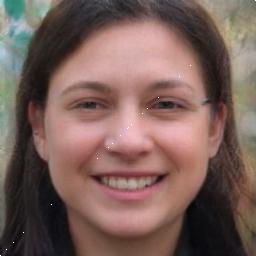}\hfill
    \includegraphics[width=0.142\linewidth]{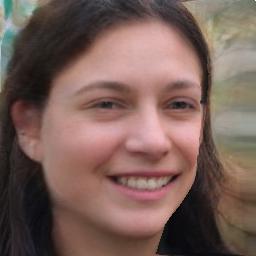}\hfill
    \includegraphics[width=0.142\linewidth]{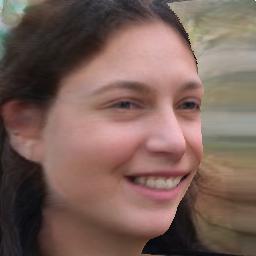}\hfill
    \includegraphics[width=0.142\linewidth]{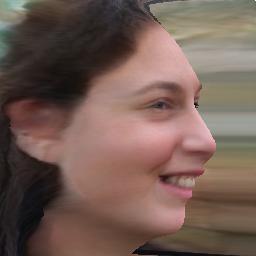}
\end{minipage}}\\
\vspace{-1.0em}
\caption{Qualitative comparison with state-of-the-art methods on 3D-controllable GANs. Note that all these faces are generated by randomly sampling from latent space, therefore we cannot compare the manipulation over the same face. The example faces in (c)-(f) are supposed to have a yaw degree of -60,-30,15,0,15,30,60.}
\label{fig:comparison_generation}
\end{figure}

Figure~\ref{fig:comparison_generation} shows some generation results of state-of-the-art work on 3D-controllable GANs. Szabo~\etal~\cite{szabo2019unsupervised} used a purely GAN-based model, where the discriminator is the main source of supervision. Therefore, they are able to generate realistic rendered images. However, their 3D shapes suffer from strong distortion, potentially due to the unstable nature of adversarial loss. This could be clearly observed from the side view of their 3D shapes. In comparison, our 3D shapes are realistic even from side views. Similarly, HoloGAN~\cite{szabo2019unsupervised} is another unsupervised method based on adversarial loss. But since their 3D transformation is applied implicitly in the feature space, it is not able to provide explicit shapes of the generated images, and hence it fails to generate faces of poses that are not in the original training data.  StyleRig~\cite{tewari2020stylerig}, CONFIGNet~\cite{kowalski2020config} and DiscoFaceGAN~\cite{deng2020disentangled} are three recent methods that attempt to disentangled 2D GAN with the supervision of 3DMM or other 3D engines. The difference is, StyleRig tries to disentangle a pretrained StyleGAN, as our method, while CONFIGNet and DiscoFaceGAN train a new generator from scratch with additional data generated by 3D models. Similar to HoloGAN, they only output the images but no other 3D components. From Figure~\ref{fig:comparison_generation}, it can be seen that our model, though unsupervised, is able to generalize to larger yaw degrees than most of the baselines, which are rarely seen in the training data. More results can be found in the supplementary.

\begin{figure}[t]
    \centering
    \captionsetup{font=footnotesize}
    \includegraphics[width=0.48\linewidth]{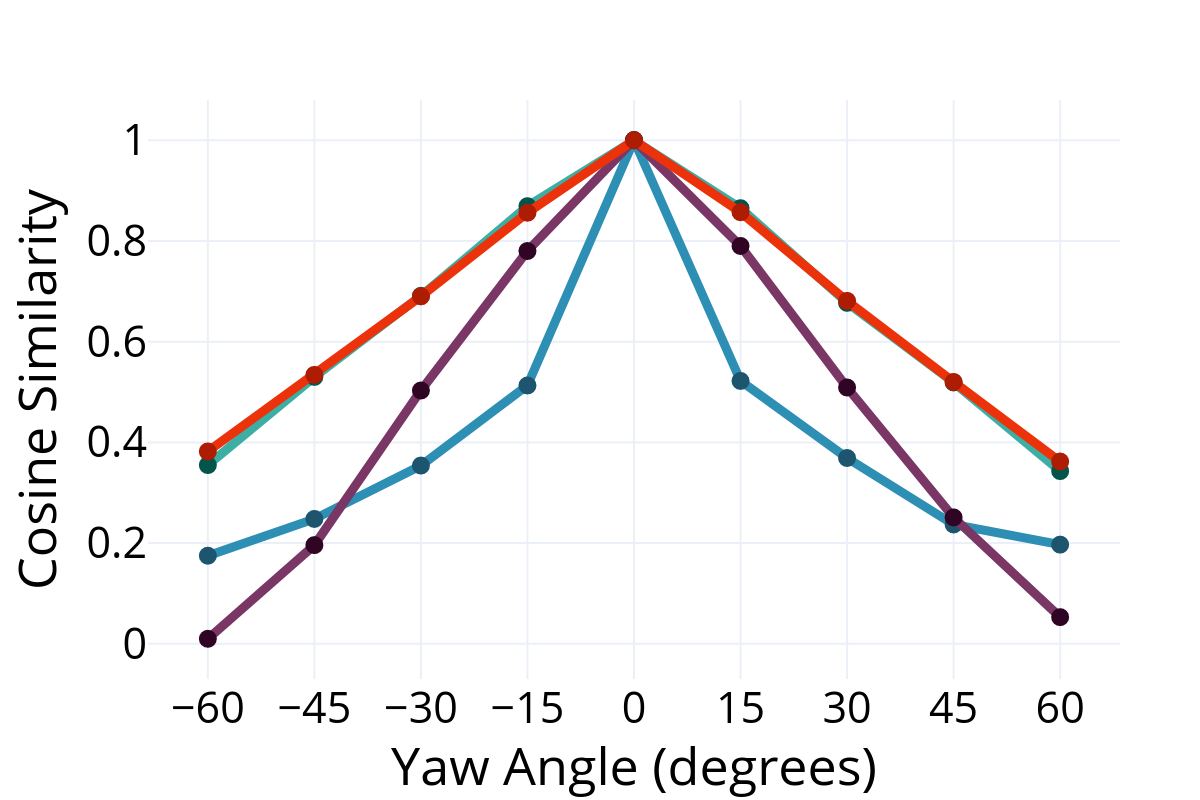}\hfill
    \includegraphics[width=0.48\linewidth]{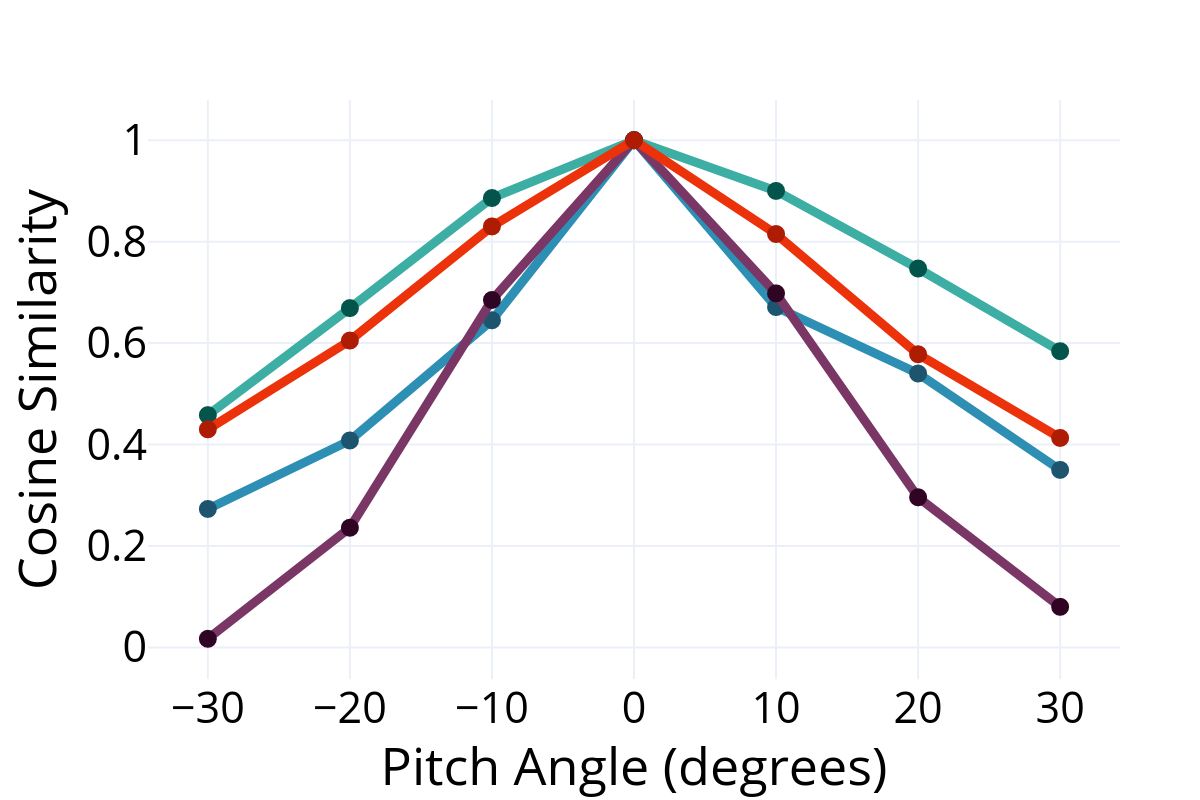}\\[0.5em]
    \includegraphics[width=0.8\linewidth]{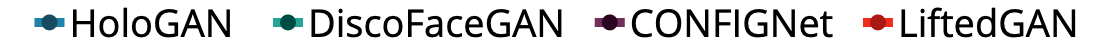}
    \caption{Quantitative comparison of the proposed LiftedGAN with state of the art works on 3D-controllable generative models. We compute the averaged identity similarity under different rotation angles using a face recognition method (ArcFace).}%
    \label{fig:HoloGAN_comparison}
\end{figure}

We further quantitatively compare our approach with three open-sourced methods: HoloGAN~\cite{nguyen2019hologan}, DiscoFaceGAN~\cite{deng2020disentangled}  and CONFIGNet~\cite{kowalski2020config} in Figure~\ref{fig:HoloGAN_comparison}. For each method, we randomly generate $1,000$ faces, each with multiple outputs under specified poses. Then a state-of-the-art face matcher, ArcFace~\cite{deng2019arcface} is used to compute the similarity between the generated frontal face and non-frontal faces. 
As can be seen in Figure~\ref{fig:HoloGAN_comparison}, in spite of the good visual quality of all methods, the identity is hardly preserved after the face rotation except DiscoFaceGAN, which is trained with the supervision of a 3DMM. In comparison, our method is able to maintain the identity information better than most methods even without any supervision.

\begin{table}[t]
\captionsetup{font=footnotesize}
\newcommand{\mr}[1]{\multirow{2}{*}{#1}}
\setlength{\tabcolsep}{4pt}
\footnotesize
\begin{center}
\begin{tabularx}{1.0\linewidth}{X c c cccc}
\toprule
\mr{Method} & \multicolumn{1}{c}{Image} &\;& \multicolumn{4}{c}{3D} \\
\cline{2-2} \cline{4-7}
& FID && Depth & Yaw & Pitch & Row\\
\midrule            
StyleGAN2 ($G_{2D})$                & 12.72 && -        & -     & - & -  \\
Wu ~\etal~\cite{wu2020unsupervised} & -     && 0.472    & 2.78  & 4.55 & 0.75  \\\hline
LiftedGAN w/o $\EL_{flip}$          & 28.69 && 0.623    & 9.37  & 5.34 & 0.91   \\
LiftedGAN w/o $\EL_{perturb}$       & 21.30 && 0.548    & 2.68  & 5.03 & 1.02    \\
LiftedGAN w/o $\EL_{idt}$           & 30.63 && 0.498    & 1.82  & 4.66 & 1.15 \\
LiftedGAN w/o $\EL_{reg_A}$         & 27.34 && 0.467    & 1.89  & 4.81 & 1.06 \\
LiftedGAN (proposed)                & 29.81 && 0.435    & 1.86  & 4.77 & 1.07    \\
\bottomrule
\end{tabularx}
\caption{Quantitative Evaluation of the ablation study. LiftedGAN refers to our method while StyleGAN2 refers to the output of base 2D GAN used for training. The 3D evluation is conducted on samples generated by StyleGAN2. The depth error is the $\ell_2$ distance between normalized depth maps and the angle errors are reported in degrees. FID represents the Fr$\acute{\text{e}}$chet inception distance.}
\label{tab:ablation}
\end{center}\vspace{-1.0em}
\end{table}

\begin{figure}
\captionsetup{font=footnotesize}
\renewcommand\twocolumn[1][]{#1}%
\newcommand{\mmc}[1]{\multicolumn{1}{c}{#1}}
\vspace{-0.8em}
\footnotesize
    \centering
    \setlength{\tabcolsep}{1pt}
    \begin{tabularx}{1.0\linewidth}{ X ccccc}
    & w/o $\EL_{flip}$ & w/o $\EL_{perturb}$ & w/o $\EL_{idt}$ & w/o $\EL_{reg_A}$ & proposed\\
    \raisebox{1.0\height}{\rotatebox[origin=c]{90}{(a) Frontal}} & 
    \includegraphics[width=0.19\linewidth]{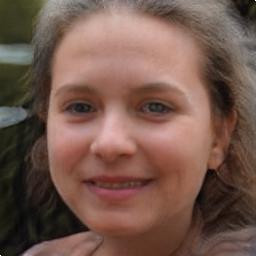} &
    \includegraphics[width=0.19\linewidth]{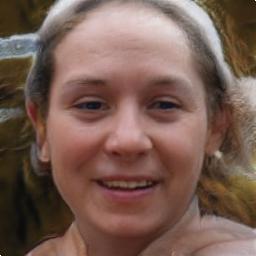} &
    \includegraphics[width=0.19\linewidth]{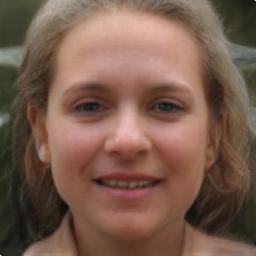} &
    \includegraphics[width=0.19\linewidth]{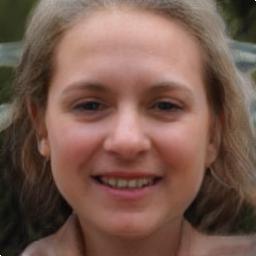} &
    \includegraphics[width=0.19\linewidth]{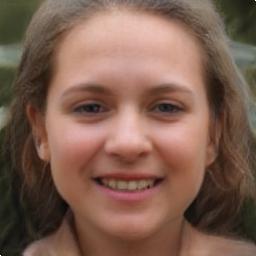} \\[-0.3em]
    \raisebox{1.0\height}{\rotatebox[origin=c]{90}{(b) Rotated}} & 
    \includegraphics[width=0.19\linewidth]{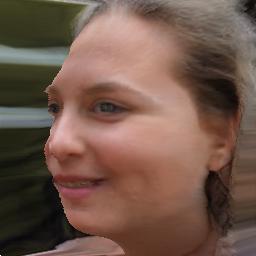} &
    \includegraphics[width=0.19\linewidth]{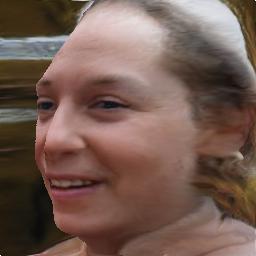} &
    \includegraphics[width=0.19\linewidth]{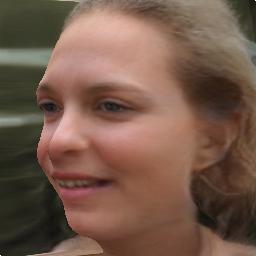} &
    \includegraphics[width=0.19\linewidth]{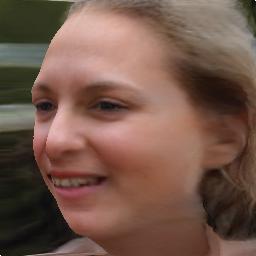} &
    \includegraphics[width=0.19\linewidth]{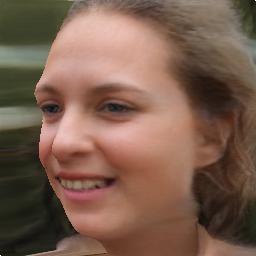} \\[-0.3em]
    \raisebox{1.2\height}{\rotatebox[origin=c]{90}{(c) Shape}} & 
    \includegraphics[width=0.19\linewidth]{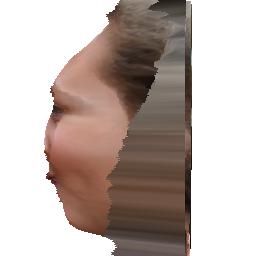} &
    \includegraphics[width=0.19\linewidth]{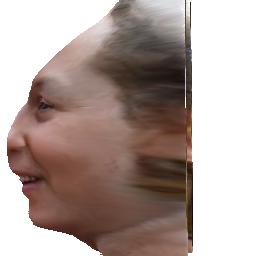} &
    \includegraphics[width=0.19\linewidth]{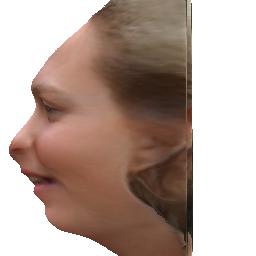} &
    \includegraphics[width=0.19\linewidth]{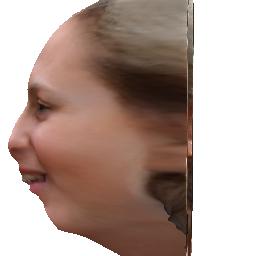} &
    \includegraphics[width=0.19\linewidth]{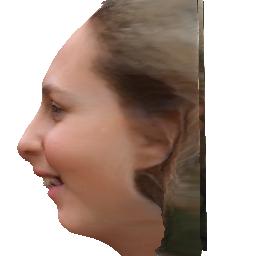} \\[-0.3em]
    \raisebox{1.0\height}{\rotatebox[origin=c]{90}{(d) Relighted}} & 
    \includegraphics[width=0.19\linewidth]{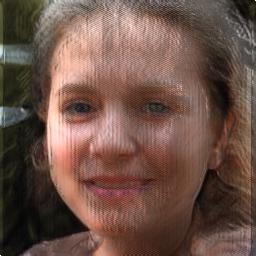} &
    \includegraphics[width=0.19\linewidth]{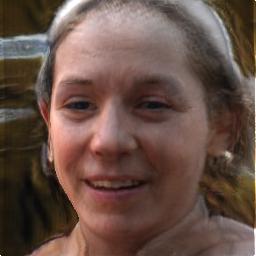} &
    \includegraphics[width=0.19\linewidth]{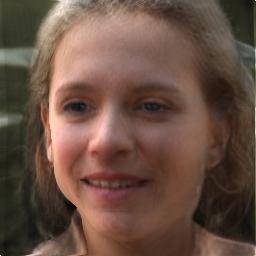} &
    \includegraphics[width=0.19\linewidth]{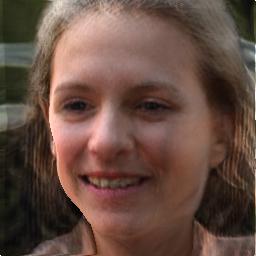} &
    \includegraphics[width=0.19\linewidth]{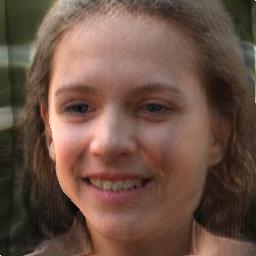} \\[-0.3em]
    \end{tabularx}
    \caption{Qualitative results of ablation study.}\vspace{-1.0em}
    \label{fig:ablation}
\end{figure}%

\subsection{Ablation Study}
\label{sec:exp:ablation}
In this section, we ablate over different loss functions to see their effect on the generated results. In particular, we remove symmetric reconstruction loss, perturbation loss, albedo consistency loss and identity regularization loss, respectively, to re-train a model for comparison. The evaluation is conducted from two perspectives. First, Fr$\acute{\text{e}}$chet inception distance (FID)~\cite{heusel2017gans} is used to evaluate the image quality, where we compare between the original 2D training data and randomly sampled images from our 3D Generator. Second, we also wish to evaluate how precise the estimated 3D shapes are. However, since our models are generative methods, for whose output we do not have ground-truth 3D annotations, it is hard to directly evaluate the 3D outputs. Therefore instead, we compute the estimated 3D shapes of $1,000$ generated images of StyleGAN2 with a state-of-the-art 3D reconstruction model~\cite{deng2019accurate} as pseudo ground-truth and compare them to the estimated shapes of our model. In detail, we evaluate both depth map error and angle prediction error. For depth map, since different camera parameters are used for our method and the estimated labels, we first normalize the depth map in terms of mean and standard deviation before computing their $\ell_2$ distance on overlapped areas. The view angles are normalized by subtracting average angle. To provide a reference, we also train a state-of-the-art unsupervised face reconstruction model~\cite{wu2020unsupervised} on the StyleGAN2 generated faces and test it on the same set of images. The quantitative and qualitative results are shown in Table~\ref{tab:ablation} and Figure~\ref{fig:ablation}, respectively. Without the symmetric reconstruction loss, we found that the model is not able to output a reasonable shape, even though it is still able to rotate the face to a certain degree. The perturbation loss is helpful for learning detailed and realistic face shapes, especially from a side view. The albedo consistency loss and identity loss has a similar effect of regularizing the output shape, as can be seen in Table~\ref{tab:ablation}. We also notice that applying the perturbation loss cause a higher FID, we believe this is caused by the identity change in the style-manipulated faces created by StyleGAN2, whose supervision removes some details on the generated 3D faces. The problem could be potentially solved by adding additional regularization loss to the StyleGAN2 manipulation output, which we leave for future work. 

\subsection{Additional Experiments}

\begin{figure}
\captionsetup{font=footnotesize}
\renewcommand{\arraystretch}{0.1}
\centering
\small
\setlength\tabcolsep{0px}
\newcommand{\www}{0.235\linewidth}
\newcolumntype{Y}{>{\centering\arraybackslash}X}
    \begin{tabularx}{\linewidth}{Y>{\centering\arraybackslash}cccc}
    & Input & Pose 1 & Pose 2 & Pose 3 \\
    \raisebox{1.0\height}{\rotatebox[origin=c]{90}{(a) Original}}\; & 
    \includegraphics[width=\www]{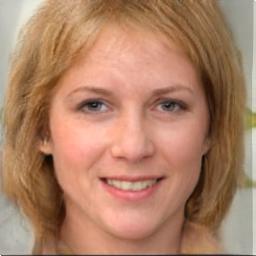}&
    \includegraphics[width=\www]{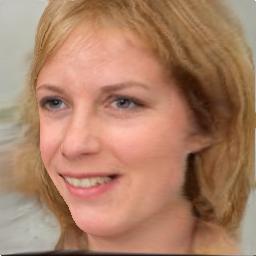}&
    \includegraphics[width=\www]{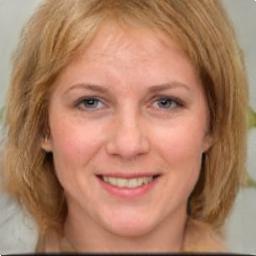}&
    \includegraphics[width=\www]{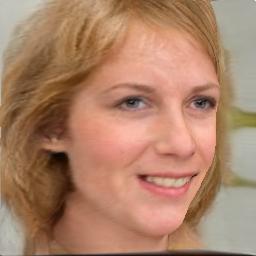}\\
    \raisebox{1.2\height}{\rotatebox[origin=c]{90}{(b) Style 1}}\; & 
    \includegraphics[width=\www]{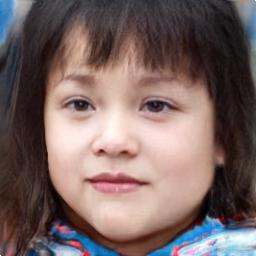}&
    \includegraphics[width=\www]{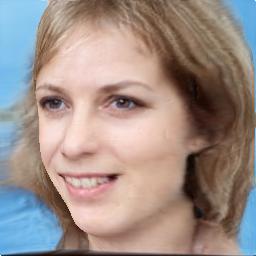}&
    \includegraphics[width=\www]{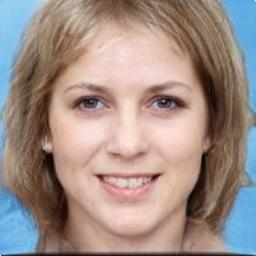}&
    \includegraphics[width=\www]{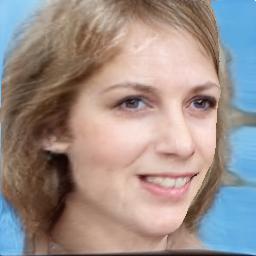}\\
    \raisebox{1.2\height}{\rotatebox[origin=c]{90}{(c) Style 2}}\; & 
    \includegraphics[width=\www]{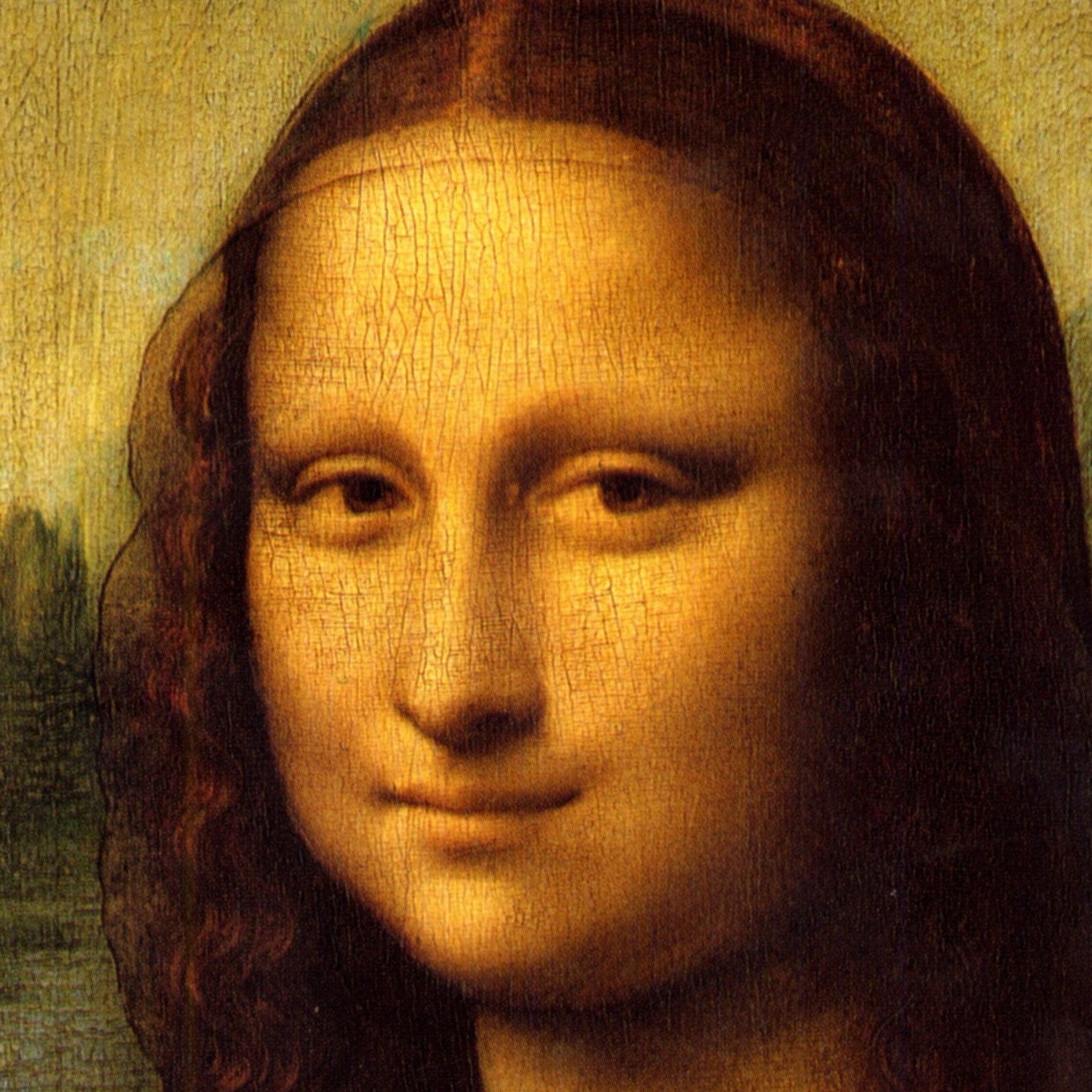}&
    \includegraphics[width=\www]{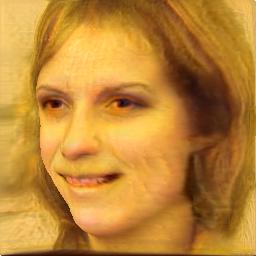}&
    \includegraphics[width=\www]{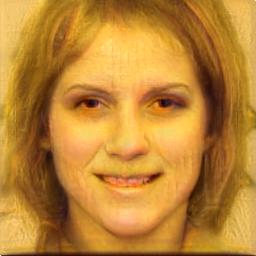}&
    \includegraphics[width=\www]{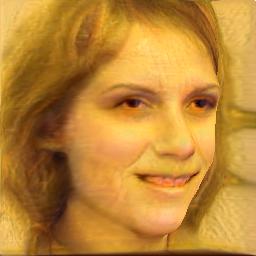}\\
    \raisebox{1.2\height}{\rotatebox[origin=c]{90}{(d) Style 3}}\; & 
    \includegraphics[width=\www,height=\www]{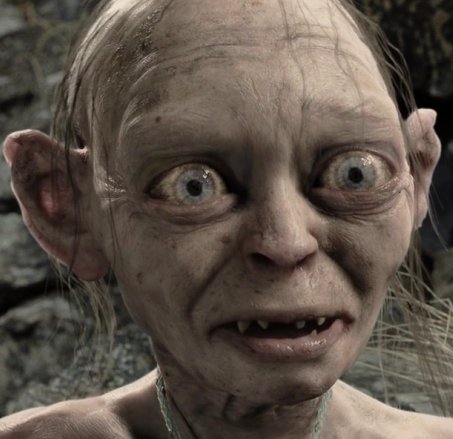}&
    \includegraphics[width=\www]{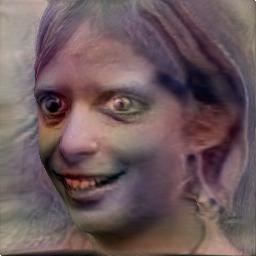}&
    \includegraphics[width=\www]{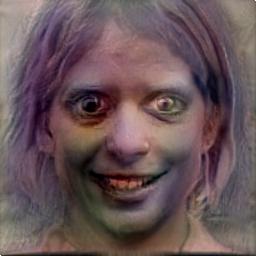}&
    \includegraphics[width=\www]{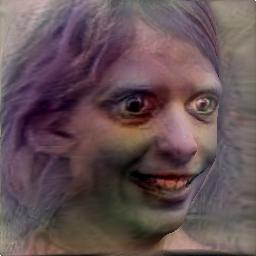}\\[-0.2em]
    \end{tabularx}
    \caption{Example results of style transfer. Row (a) shows a randomly sampled image and its different views. Rows (b)-(d) show the results of transferring the style to images in row (a) while changing the pose of the generated face. The first column in rows (b)-(d) shows the input images used to extract the style. The input images are first embedded into the latent space of StyleGAN2 by optimizing the latent code, then intermediate features are combined to change the texture as in~\cite{stylegan2}.}
    \label{fig:generation_style_transfer}
\end{figure}

\paragraph{Structure vs. Style }The original StyleGAN2 is famous for being able to control and transfer the style of generated images~\cite{stylegan,stylegan2}. However, this notion of ``style" is rather ambiguous: it could mean coarse structure (e.g. pose), fine-grained structure (e.g. expression), or color style. This is because the style control is learned in an unsupervised manner by AdaIN modules~\cite{huang2017arbitrary} at different layers. Since our work explicitly disentangles the 3D information, it implies that the 3D geometry and other styles could be decoupled in our generator. Figure~\ref{fig:generation_style_transfer} shows a few examples of transferring different styles to a generated face while controlling the pose of that face. It could be seen that style transfer only affects the texture color and local structures, while the overall structure of the image is consistent across different poses, as controlled by the user.

\begin{figure}[t]
\captionsetup{font=footnotesize}
\centering
\footnotesize
\setlength\tabcolsep{1px}
\newcommand{\www}{0.19\linewidth}
\renewcommand{\arraystretch}{0.1}
\newcolumntype{Y}{>{\centering\arraybackslash}X}
\begin{tabularx}{\linewidth}{Y>{\centering\arraybackslash}c}
    \raisebox{1.0\height}{\rotatebox[origin=c]{90}{StyleGAN2}} & 
    \includegraphics[width=\www]{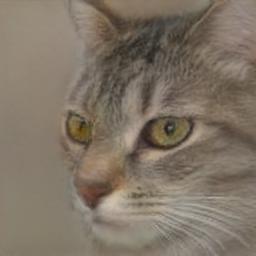}\hfill
    \includegraphics[width=\www]{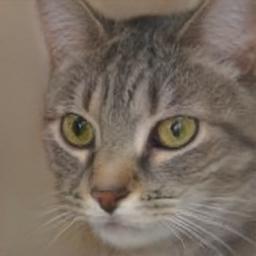}\hfill
    \includegraphics[width=\www]{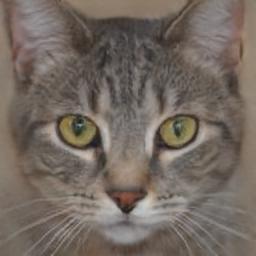}\hfill
    \includegraphics[width=\www]{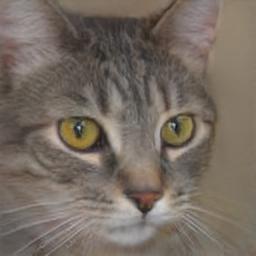}\hfill
    \includegraphics[width=\www]{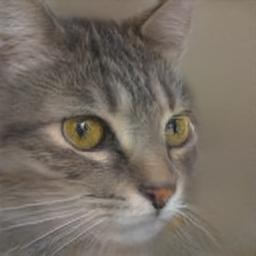}\\
    \raisebox{0.8\height}{\rotatebox[origin=c]{90}{3D Generator}} & 
    \includegraphics[width=\www]{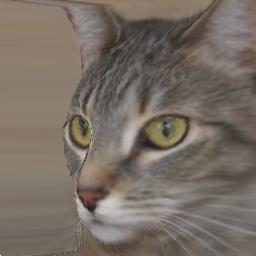}\hfill
    \includegraphics[width=\www]{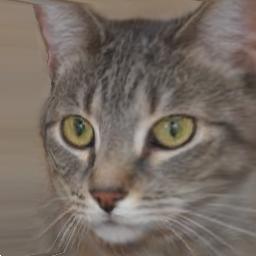}\hfill
    \includegraphics[width=\www]{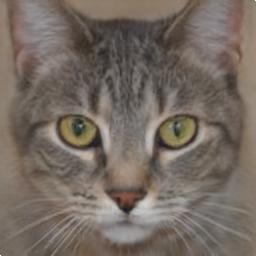}\hfill
    \includegraphics[width=\www]{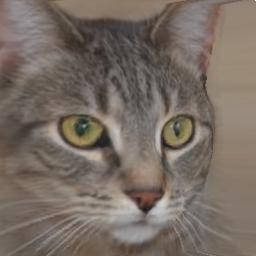}\hfill
    \includegraphics[width=\www]{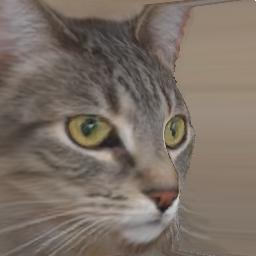}\\
    \raisebox{1.5\height}{\rotatebox[origin=c]{90}{Shape}} & 
    \includegraphics[width=\www]{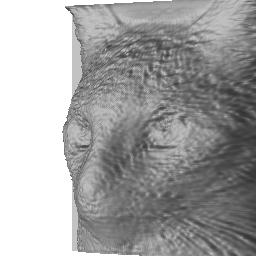}\hfill
    \includegraphics[width=\www]{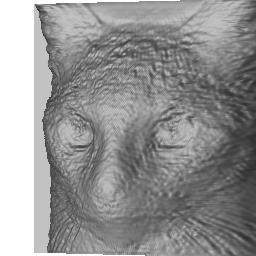}\hfill
    \includegraphics[width=\www]{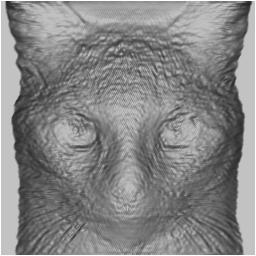}\hfill
    \includegraphics[width=\www]{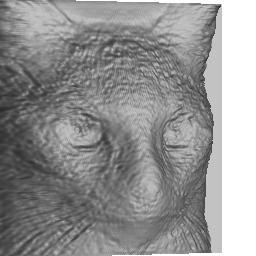}\hfill
    \includegraphics[width=\www]{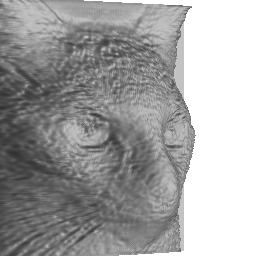}\\
\end{tabularx}\vspace{-0.5em}
    \caption{Results of controlled generation on cat heads.}
    \label{fig:cats}\vspace{-1.5em}
\end{figure}

\paragraph{Other Objects} Except the identity preservation loss, our method is designed to be object-agnostic. So it would be interesting to see whether our method works on other objects as well. However, as shown in Section~\ref{sec:exp:ablation}, symmetric constraint still plays a important role in the current method. Thus, we conduct the experiment on another symmetric object, cat face~\cite{catface}, with the same settings except dropping the identity preservation loss.  As shown in Figure~\ref{fig:cats}, the proposed method generalizes to other symmetric objects in terms of both controlling StyleGAN2 images and learning an additional 3D generator.

\section{Conclusions}
We propose a method to distill a pre-trained StyleGAN2 into a 3D-aware generative model, called LiftedGAN. LiftedGAN disentangles the latent space of StyleGAN2 into pose, light, texture and a depth map from which face images can be generated with a differentiable renderer. During training, the 2D StyleGAN2 is used as a teacher network to generate proxy images to guide the learning of the 3D generator. The learning process is completely self-supervised, i.e. it neither requires 3D supervision nor 3D morphable model. During inference, the 3D generator is able to generate face images of selected pose angles and lighting conditions. Compared with 3D-controllable generative models that are based on feature manipulation, our model gives explicit shape information of generated faces and hence preserves the content better during face rotation. Compared with 3D generative models, our model provides more realistic face shapes by distilling the knowledge from StyleGAN2. Potential future includes: (1) combining our method with the training of StyleGAN2 to obtain a more disentangled 2D generator and (2) extending the method to more object domains.

{\small
\bibliographystyle{ieee_fullname}
\bibliography{egbib}
}

\clearpage
\section*{Appendix}
\appendix

\section{Joint Probability as Lower Bound}
Here we show the relationship between $\log p(I'_w,w')$ and the likelihood $p(I'_w)=\int_w p_{G_{2D}}(I'_w|w)p(w) dw$. In particular, by introducing an additional encoder $q(w)$, the variational lower bound~\cite{kingma2014auto} is given by:
\begin{equation}
\label{eq:ELBO}
    \log p(I'_w) \ge \mathbb{E}_{w\sim q(w)}[\log \frac{p(I'_w,w)}{q(w)}].
\end{equation}
In the original Variational AutoEncoder~\cite{kingma2014auto}, $q$ should be a conditional distribution defined as an image encoder to approximate the posterior $p(w|I'_w)$. Here, we use the manipulation network to replace the image encoder. Formally, consider $q$ to be a Gaussian distribution conditioned on the style code $\hat{w}$, perturbation parameters $V'$ and $L'$, i.e. $q(w|M(\hat{w},V',L'))=q(w|w')=\mathcal{N}(w',\sigma^2_{w'}\mathbf{I})$, if we further consider $q$ to be a deterministic approximation, i.e. $\sigma_{w'}$ is a fixed value and $\sigma_{w'}\rightarrow 0$, the lower bound in Equation~\ref{eq:ELBO} becomes $\log p(I'_w,w')+c$, where $c$ is a constant and could be omitted.

\section{Additional Implementation Details}
We use an Adam optimizer~\cite{kingma2014adam} with a learning rate of 1e-4 to train the model. The 3D generator is trained for $30,000$ steps. For the perceptual loss, we average the $\ell_2$ distance between the output of \texttt{relu2\_2},  \texttt{relu3\_3}, \texttt{relu4\_3} to compute the loss. The feature maps are $\ell_2$-normalized on each pixel. Further, the images are downsampled with a scale of $\times1$, $\times2$, $\times4$ and the perceptual losses are averaged for different resolutions. For the 3D renderer, we assume a fov of 10. The output depth maps are normalized between $(0.9,1.1)$. The output viewpoint rotation and translation are normalized between (-$60^{\circ}$,$60^{\circ}$) and (-0.1,0.1), respectively. The lighting coefficients are normalized to (0,1). For the perturbation, we empirically sample yaw angles from a uniform distribution over [$-45^{\circ}$,$45^{\circ}$], and pitch angles from [$-10^{\circ}$,$10^{\circ}$]. The roll angle is fixed at 0. For the lighting, we found it a difficult task for StyleGAN2 to synthesize faces with manually chosen lighting parameters. Thus, we shuffle the lighting parameters across the batch. The identity regularization loss is only applied to faces that are perturbed with a yaw angle within [$-25^{\circ}$,$25^{\circ}$]. For the identity preservation loss, we train an 18-layer ResNet~\cite{deng2019arcface} as the face embedding using the CosFace loss~\cite{wang2018cosface} on the CASIA-Webface dataset~\cite{yi2014learning}.

\subsection{Network Architectures}
The viewpoint decoder $D_V$ and light decoder $D_L$ are both 4-layer MLPs with LeakyReLU~\cite{LeakyReLU} as activation function. The latent manipulation network is composed of two parts, the first part are composed of three 4-layer MLPs to encode latent code $\hat{w}$, viewpoint $V_0$ and light $L_0$, respectively, whose outputs are summed into a feature vector for the second part. The second part is another 4-layer MLP that outputs a new style code. The hidden size of all MLPs in our work is 512, same as the style code. For shape decoder $D_S$ and transformation decoder $D_T$, we mainly follow the decoder structure of Wu~\etal~\cite{wu2020unsupervised}, whose details are shown in Table~\ref{appendix:tab:shape_decoder} and Table~\ref{appendix:tab:transformation_decoder}, respectively. In detail, Conv($c_{in}$, $c_{out}$, $k$, $s$) refers to a convolutional layer with $c_{in}$ input channels, $c_{out}$ output channels, a kernel size of $k$ and a stride of $s$. Deconv is defined similarly. ``GN'' refers to the group normalization, where the argument is the number of groups. The ``InjectConv'' refers to a convolutional layer that takes the last feature map of 2D generator as input and injects its output to the decoder via summation. We found such a design helps to recover the facial details in the depth map.

\begin{table}[t]
\footnotesize
    \centering
    \begin{tabularx}{1.0\linewidth}{X c}
    \toprule
    Shape Decoder                           & Output size \\
    \midrule
    4-layer MLP                             & 512 \\
    Deconv(512,512,4,1) + ReLU                & 512$\times$4$\times$4 \\
    Conv(512,512,3,1) + ReLU                  & 512$\times$4$\times$4 \\
    Deconv(512,256,4,2) + GN(64) + ReLU         & 256$\times$8$\times$8 \\
    Conv(256,256,3,1) + GN(64) + ReLU           & 256$\times$8$\times$8 \\
    Deconv(256,128,4,2) + GN(32) + ReLU         & 128$\times$16$\times$16 \\
    Conv(128,128,3,1) + GN(32) + ReLU           & 128$\times$16$\times$16 \\
    Deconv(128,64,4,2) + GN(16) + ReLU          & 64$\times$32$\times$32 \\
    Conv(64,64,3,1) + GN(16) + ReLU             & 64$\times$32$\times$32 \\
    Deconv(64,32,4,2) + GN(8) + ReLU            & 32$\times$64$\times$64 \\
    Conv(32,32,3,1) + GN(8) + ReLU              & 32$\times$64$\times$64 \\
    Deconv(32,16,4,2) + GN(4) + ReLU            & 16$\times$128$\times$128 \\
    Conv(16,16,3,1) + GN(4) + ReLU              & 16$\times$128$\times$128 \\
    Upsample(2)  +  InjectConv(32,161,1)        & 16$\times$128$\times$128 \\
    Conv(16,16,3,1) + GN(4) + ReLU              & 16$\times$256$\times$256 \\
    Conv(16,16,5,1) + GN(4) + ReLU              & 16$\times$256$\times$256 \\
    Conv(16,1,5,1)                          & 1$\times$256$\times$256 \\
    \bottomrule
    \end{tabularx}
    \caption{The architecture of shape decoder.}
    \label{appendix:tab:shape_decoder}
\end{table}

\begin{table}[t]
\footnotesize
    \centering
    \begin{tabularx}{1.0\linewidth}{X c}
    \toprule
    Transformation Map Decoder              & Output size \\
    \midrule
    4-layer MLP                             & 512 \\
    Deconv(512,512,4,1) + ReLU                & 512$\times$4$\times$4 \\
    Deconv(512,256,4,2) + GN(64) + ReLU         & 256$\times$8$\times$8 \\
    Deconv(256,128,4,2) + GN(32) + ReLU         & 128$\times$16$\times$16 \\
    Deconv(128,64,4,2) + GN(16) + ReLU          & 64$\times$32$\times$32 \\
    Deconv(64,32,4,2) + GN(8) + ReLU            & 32$\times$64$\times$64 \\
    Deconv(32,16,4,2) + GN(4) + ReLU            & 16$\times$128$\times$128 \\
    Upsample(2) + Conv(16,16,3,1) + GN(4) + ReLU     & 16$\times$256$\times$256 \\
    Conv(16,1,5,1)                          & 1$\times$256$\times$256 \\
    \bottomrule
    \end{tabularx}
    \caption{The architecture of transformation map decoder.}
    \label{appendix:tab:transformation_decoder}
\end{table}

\subsection{Additional Results}
In Figure~\ref{appendix:fig:comparison_generation_yaw}, we show more results of our method and other baselines for rotating faces into different yaw angles. Figure~\ref{appendix:fig:comparison_generation_pitch} shows the results of rotating faces into different pitch angles. Note that for HoloGAN~~\cite{nguyen2019hologan}, we use the officially released code and train the model on our dataset. For CONFIGNet~\cite{kowalski2020config} and DiscoFaceGAN~\cite{deng2020disentangled}, since they involve additional synthesized data for training,  we use their pre-trained models. Although most baselines are able to generate high-quality face images under near-frontal poses, clear content change could be observed in larger poses. In particular, HoloGAN and CONFIGNet generates blurred faces for larger poses, while the expression changes in the results of DiscoFaceGAN. In comparison, our method, though also suffer from quality degradation in larger poses, has a stricter control in terms of the content. We also provide the FID score of different methods in Table~\ref{appendix:tab:comparison_fid}, where our method achieves second best image quality, even though our images are re-rendered from 2D generated images.

In Figure~\ref{appendix:fig:generation_multiview_yaw} and Figure~\ref{appendix:fig:generation_multiview_pitch}, we show additional examples of rotating generated faces with different yaw and pitch angles, respectively. In brief, we observe that the pre-trained StyleGAN2 (with our style manipulation network) is able to change poses to a certain degree, but fails to generate faces with larger pose angles that do not exist in the training data. Further, we see a similar trend of expression change when changing pitch angles as in DiscoFaceGAN, possibly due to the intrinsic bias in the FFHQ dataset. In contrast, the 3D generator distilled from StyleGAN2 extends this rotation capability to a larger degree with better content preservation.

\begin{table}[t]
\captionsetup{font=footnotesize}
\newcommand{\mr}[1]{\multirow{2}{*}{#1}}
\setlength{\tabcolsep}{8pt}
\footnotesize
\begin{center}
\begin{tabularx}{0.7\linewidth}{X c}
\toprule
Method & FID$\downarrow$ \\
\midrule
HoloGAN~\etal\cite{nguyen2019hologan}           & 100.28 \\
DiscoFaceGAN~\etal\cite{deng2020disentangled}   & 12.90 \\
CONFIGNet~\etal\cite{kowalski2020config}        & 43.05 \\\hline
LiftedGAN (proposed)                            & 29.81 \\
\bottomrule
\end{tabularx}
\caption{Quantitative Evaluation of the generated image quality between this work and baselines. The scores of ``DiscoFaceGAN'' and ``CONFIGNet'' are reported in their original papers. The scores of HoloGAN and our method are computed from sampled images of random poses.}
\label{appendix:tab:comparison_fid}
\end{center}
\end{table}

\clearpage

\begin{figure*}[t]
\captionsetup{font=small}
\centering
\footnotesize
\setlength\tabcolsep{1px}
\newcommand{\www}{0.108\linewidth}
\renewcommand{\arraystretch}{0.2}
\newcolumntype{Y}{>{\centering\arraybackslash}X}
\begin{tabularx}{1.0\linewidth}{Y>{\centering\arraybackslash}c}
    \raisebox{1.2\height}{\rotatebox[origin=c]{90}{HoloGAN}} &
    \includegraphics[width=\www]{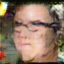}\hfill
    \includegraphics[width=\www]{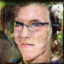}\hfill
    \includegraphics[width=\www]{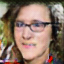}\hfill
    \includegraphics[width=\www]{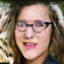}\hfill
    \includegraphics[width=\www]{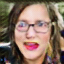}\hfill
    \includegraphics[width=\www]{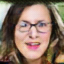}\hfill
    \includegraphics[width=\www]{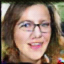}\hfill
    \includegraphics[width=\www]{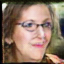}\hfill
    \includegraphics[width=\www]{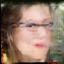}\\
    \raisebox{0.9\height}{\rotatebox[origin=c]{90}{DiscoFaceGAN}} &
    \includegraphics[width=\www]{fig/comparison_generation/DiscoFaceGAN_yaw/001/00.jpg}\hfill
    \includegraphics[width=\www]{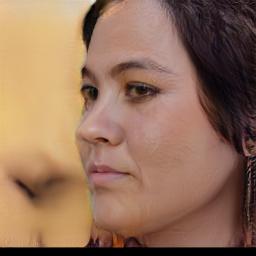}\hfill
    \includegraphics[width=\www]{fig/comparison_generation/DiscoFaceGAN_yaw/001/02.jpg}\hfill
    \includegraphics[width=\www]{fig/comparison_generation/DiscoFaceGAN_yaw/001/03.jpg}\hfill
    \includegraphics[width=\www]{fig/comparison_generation/DiscoFaceGAN_yaw/001/04.jpg}\hfill
    \includegraphics[width=\www]{fig/comparison_generation/DiscoFaceGAN_yaw/001/05.jpg}\hfill
    \includegraphics[width=\www]{fig/comparison_generation/DiscoFaceGAN_yaw/001/06.jpg}\hfill
    \includegraphics[width=\www]{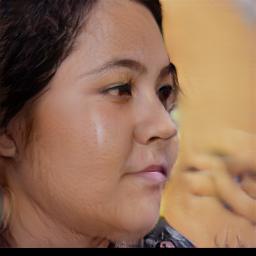}\hfill
    \includegraphics[width=\www]{fig/comparison_generation/DiscoFaceGAN_yaw/001/08.jpg}\\
    \raisebox{1.0\height}{\rotatebox[origin=c]{90}{CONFIGNet}} &
    \includegraphics[width=\www]{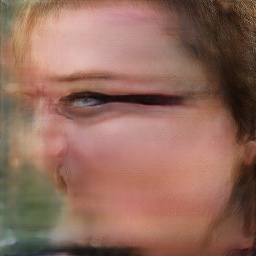}\hfill
    \includegraphics[width=\www]{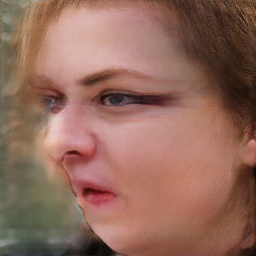}\hfill
    \includegraphics[width=\www]{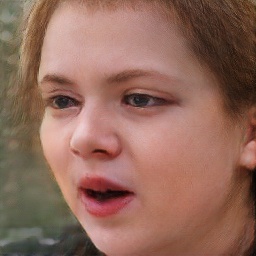}\hfill
    \includegraphics[width=\www]{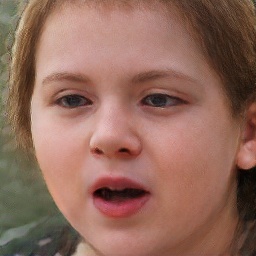}\hfill
    \includegraphics[width=\www]{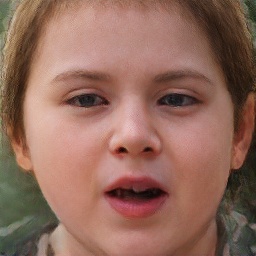}\hfill
    \includegraphics[width=\www]{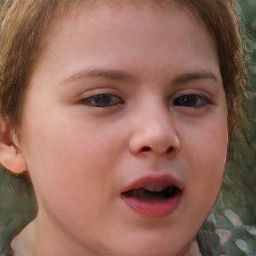}\hfill
    \includegraphics[width=\www]{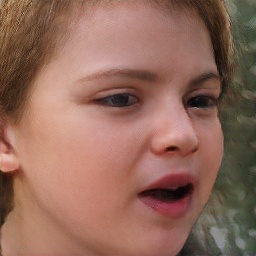}\hfill
    \includegraphics[width=\www]{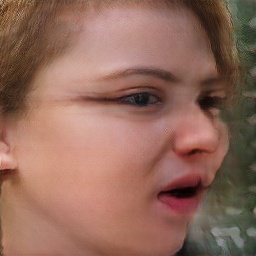}\hfill
    \includegraphics[width=\www]{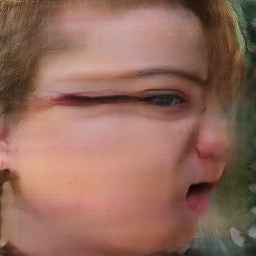}\\
    \raisebox{2.5\height}{\rotatebox[origin=c]{90}{Ours}} &
    \includegraphics[width=\www]{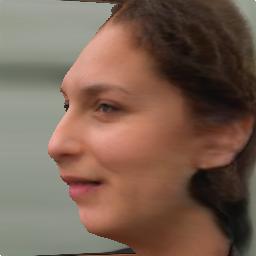}\hfill
    \includegraphics[width=\www]{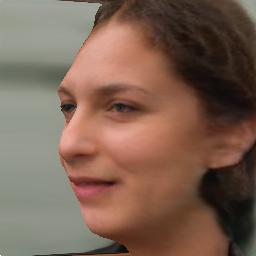}\hfill
    \includegraphics[width=\www]{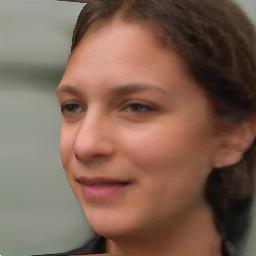}\hfill
    \includegraphics[width=\www]{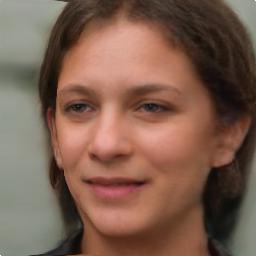}\hfill
    \includegraphics[width=\www]{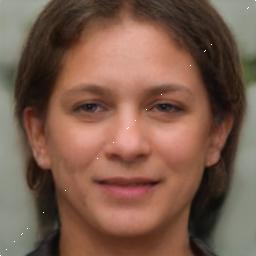}\hfill
    \includegraphics[width=\www]{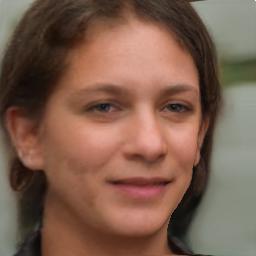}\hfill
    \includegraphics[width=\www]{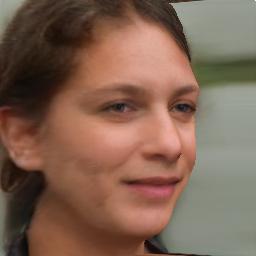}\hfill
    \includegraphics[width=\www]{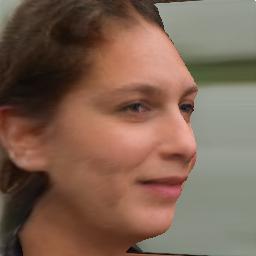}\hfill
    \includegraphics[width=\www]{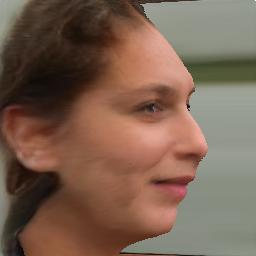}\\
    \raisebox{2.5\height}{\rotatebox[origin=c]{90}{Ours}} &
    \includegraphics[width=\www]{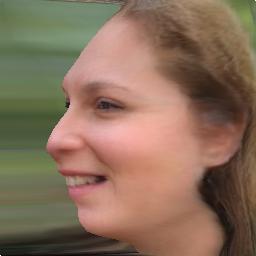}\hfill
    \includegraphics[width=\www]{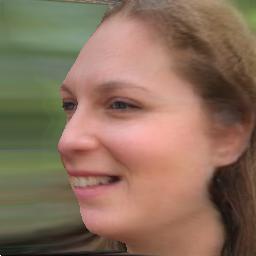}\hfill
    \includegraphics[width=\www]{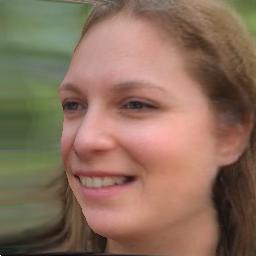}\hfill
    \includegraphics[width=\www]{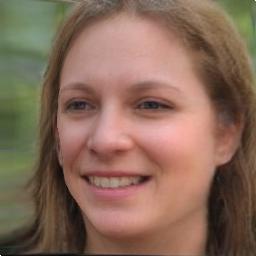}\hfill
    \includegraphics[width=\www]{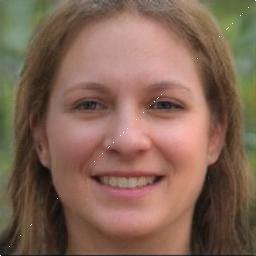}\hfill
    \includegraphics[width=\www]{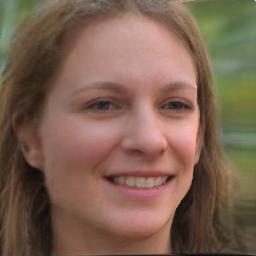}\hfill
    \includegraphics[width=\www]{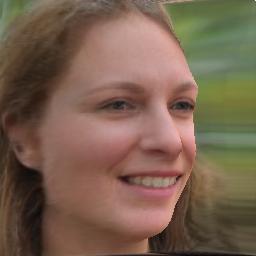}\hfill
    \includegraphics[width=\www]{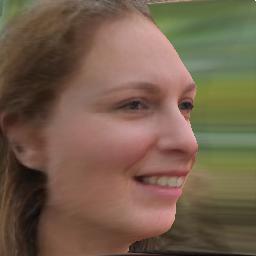}\hfill
    \includegraphics[width=\www]{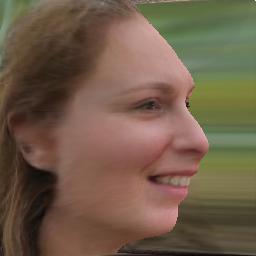}\\
\end{tabularx}
\caption{Qualitative comparison with state-of-the-art methods on 3D-controllable GANs. Note that all these faces are generated by randomly sampling from latent space, therefore we cannot compare the manipulation over the same face. The example faces are supposed to have a yaw degree of -60,-45,-30,15,0,15,30,45,60.}
\label{appendix:fig:comparison_generation_yaw}
\end{figure*}

\begin{figure*}[t]
\captionsetup{font=small}
\centering
\footnotesize
\setlength\tabcolsep{1px}
\newcommand{\www}{0.108\linewidth}
\renewcommand{\arraystretch}{0.2}
\newcolumntype{Y}{>{\centering\arraybackslash}X}
\begin{tabularx}{0.8\linewidth}{Y>{\centering\arraybackslash}c}
    \raisebox{1.2\height}{\rotatebox[origin=c]{90}{HoloGAN}} &
    \includegraphics[width=\www]{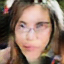}\hfill
    \includegraphics[width=\www]{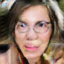}\hfill
    \includegraphics[width=\www]{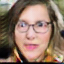}\hfill
    \includegraphics[width=\www]{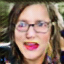}\hfill
    \includegraphics[width=\www]{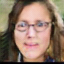}\hfill
    \includegraphics[width=\www]{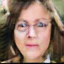}\hfill
    \includegraphics[width=\www]{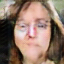}\\
    \raisebox{0.9\height}{\rotatebox[origin=c]{90}{DiscoFaceGAN}} &
    \includegraphics[width=\www]{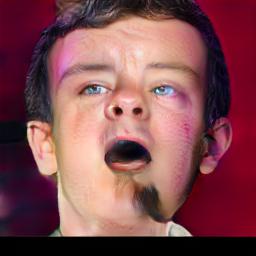}\hfill
    \includegraphics[width=\www]{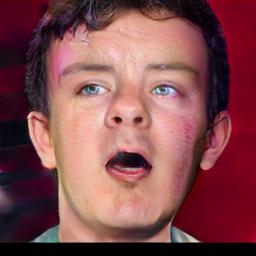}\hfill
    \includegraphics[width=\www]{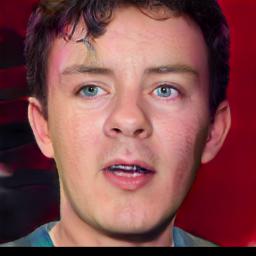}\hfill
    \includegraphics[width=\www]{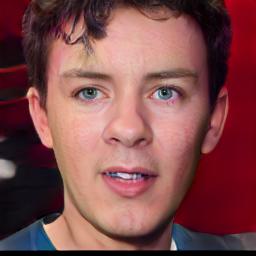}\hfill
    \includegraphics[width=\www]{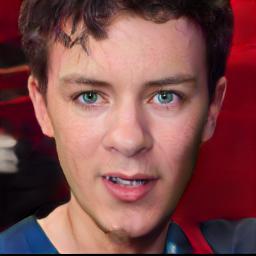}\hfill
    \includegraphics[width=\www]{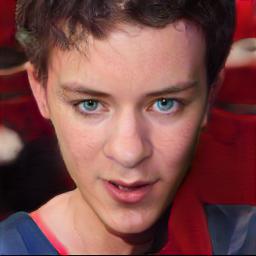}\hfill
    \includegraphics[width=\www]{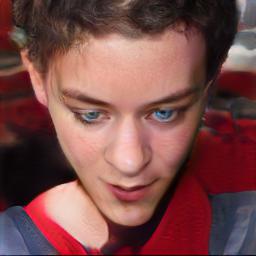}\\
    \raisebox{1.0\height}{\rotatebox[origin=c]{90}{CONFIGNet}} &
    \includegraphics[width=\www]{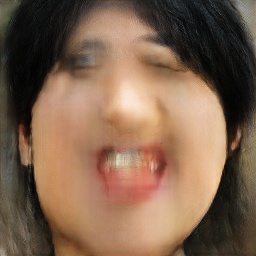}\hfill
    \includegraphics[width=\www]{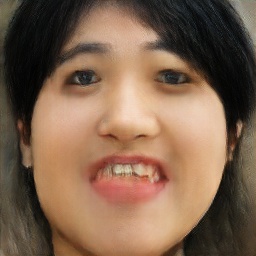}\hfill
    \includegraphics[width=\www]{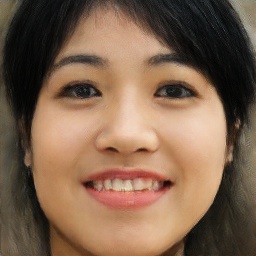}\hfill
    \includegraphics[width=\www]{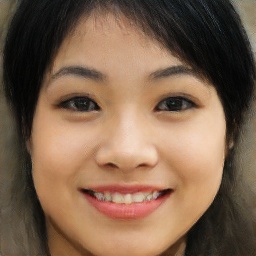}\hfill
    \includegraphics[width=\www]{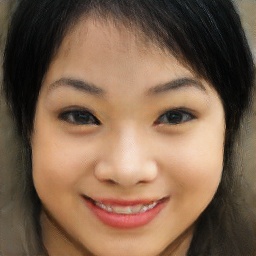}\hfill
    \includegraphics[width=\www]{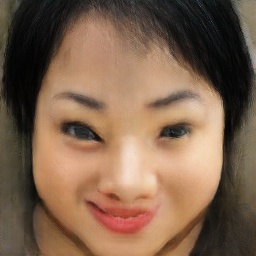}\hfill
    \includegraphics[width=\www]{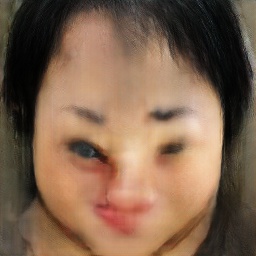}\\
    \raisebox{2.5\height}{\rotatebox[origin=c]{90}{Ours}} &
    \includegraphics[width=\www]{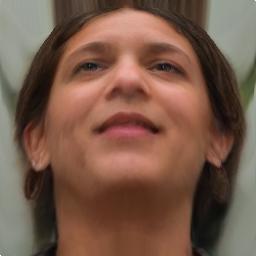}\hfill
    \includegraphics[width=\www]{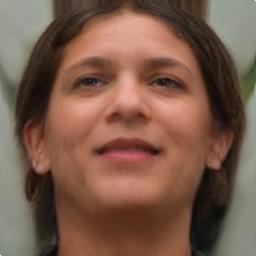}\hfill
    \includegraphics[width=\www]{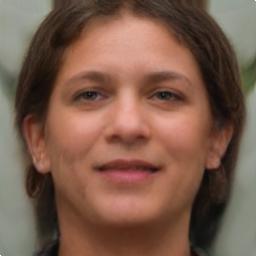}\hfill
    \includegraphics[width=\www]{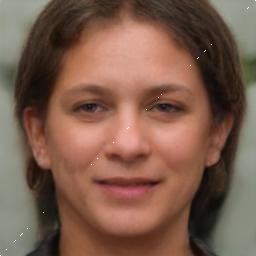}\hfill
    \includegraphics[width=\www]{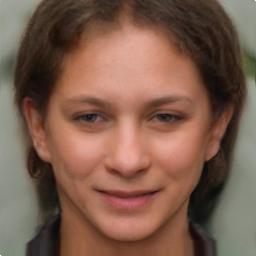}\hfill
    \includegraphics[width=\www]{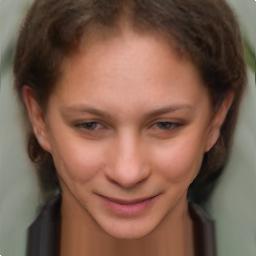}\hfill
    \includegraphics[width=\www]{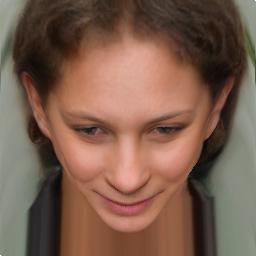}\\
    \raisebox{2.5\height}{\rotatebox[origin=c]{90}{Ours}} &
    \includegraphics[width=\www]{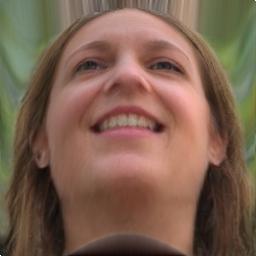}\hfill
    \includegraphics[width=\www]{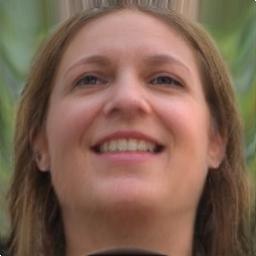}\hfill
    \includegraphics[width=\www]{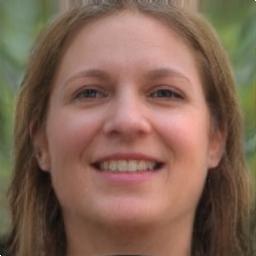}\hfill
    \includegraphics[width=\www]{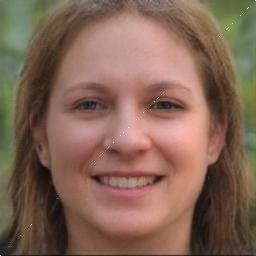}\hfill
    \includegraphics[width=\www]{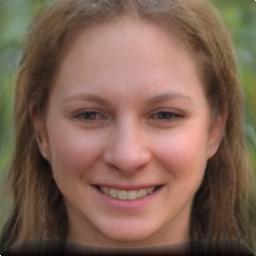}\hfill
    \includegraphics[width=\www]{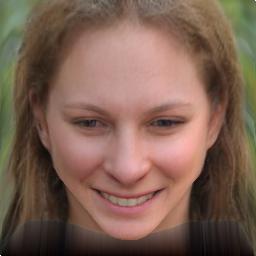}\hfill
    \includegraphics[width=\www]{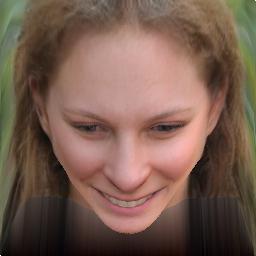}\\
\end{tabularx}
\caption{Qualitative comparison with state-of-the-art methods on 3D-controllable GANs. Note that all these faces are generated by randomly sampling from latent space, therefore we cannot compare the manipulation over the same face. The example faces are supposed to have a pitch degree of -30,-20,-10,0,10,20,30.}
\label{appendix:fig:comparison_generation_pitch}
\end{figure*}

\begin{figure*}[t]
\captionsetup{font=small}
\centering
\footnotesize
\setlength\tabcolsep{1px}
\newcommand{\www}{0.108\linewidth}
\renewcommand{\arraystretch}{0.0}
\newcolumntype{Y}{>{\centering\arraybackslash}X}
\begin{tabularx}{\linewidth}{Y>{\centering\arraybackslash}c}
    \raisebox{1.2\height}{\rotatebox[origin=c]{90}{StyleGAN2}} &
    \includegraphics[width=\www]{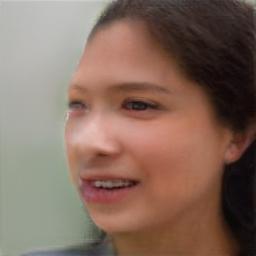}\hfill
    \includegraphics[width=\www]{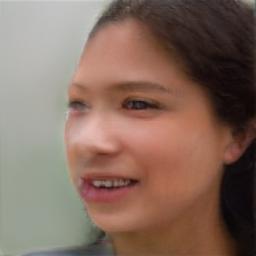}\hfill
    \includegraphics[width=\www]{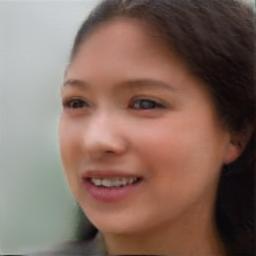}\hfill
    \includegraphics[width=\www]{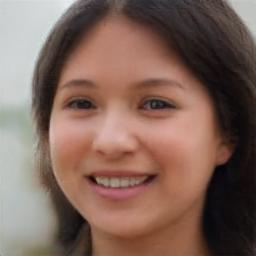}\hfill
    \includegraphics[width=\www]{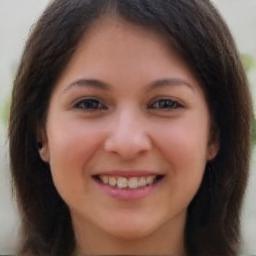}\hfill
    \includegraphics[width=\www]{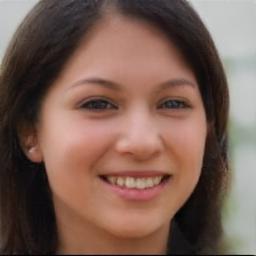}\hfill
    \includegraphics[width=\www]{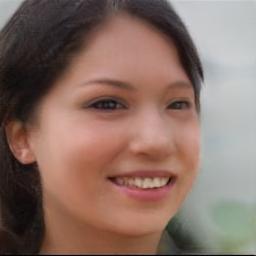}\hfill
    \includegraphics[width=\www]{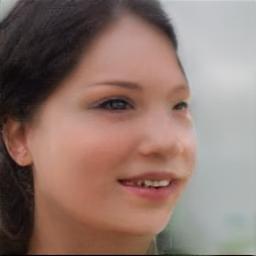}\hfill
    \includegraphics[width=\www]{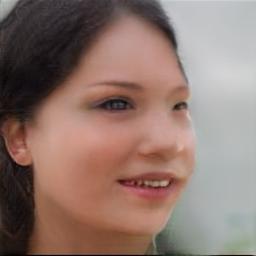}\\
    \raisebox{1.5\height}{\rotatebox[origin=c]{90}{Rendered}} &
    \includegraphics[width=\www]{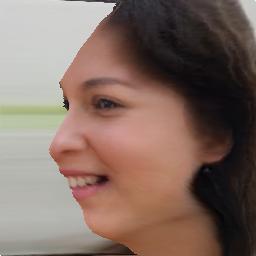}\hfill
    \includegraphics[width=\www]{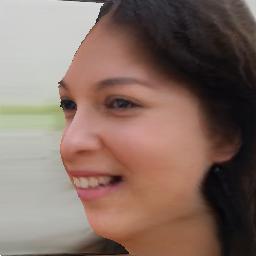}\hfill
    \includegraphics[width=\www]{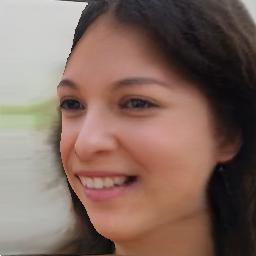}\hfill
    \includegraphics[width=\www]{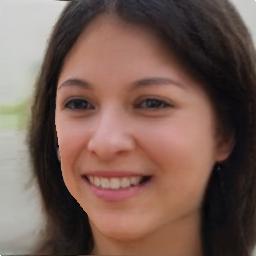}\hfill
    \includegraphics[width=\www]{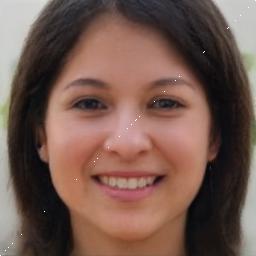}\hfill
    \includegraphics[width=\www]{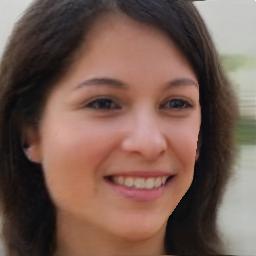}\hfill
    \includegraphics[width=\www]{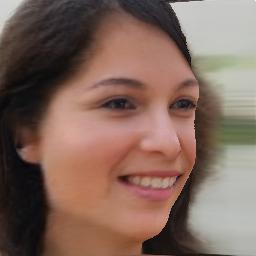}\hfill
    \includegraphics[width=\www]{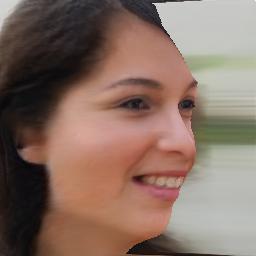}\hfill
    \includegraphics[width=\www]{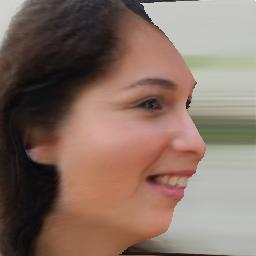}\\[0.5em]
    \raisebox{1.2\height}{\rotatebox[origin=c]{90}{StyleGAN2}} &
    \includegraphics[width=\www]{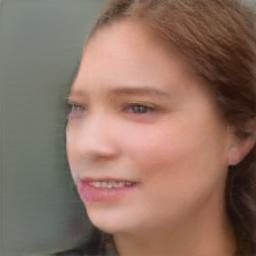}\hfill
    \includegraphics[width=\www]{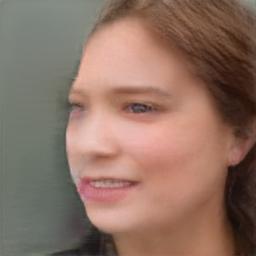}\hfill
    \includegraphics[width=\www]{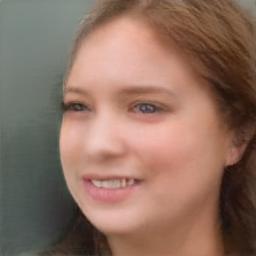}\hfill
    \includegraphics[width=\www]{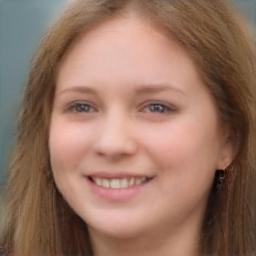}\hfill
    \includegraphics[width=\www]{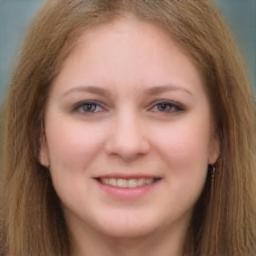}\hfill
    \includegraphics[width=\www]{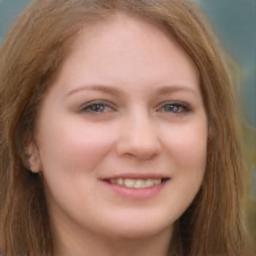}\hfill
    \includegraphics[width=\www]{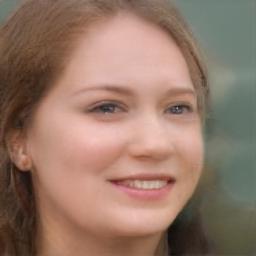}\hfill
    \includegraphics[width=\www]{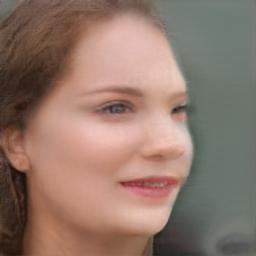}\hfill
    \includegraphics[width=\www]{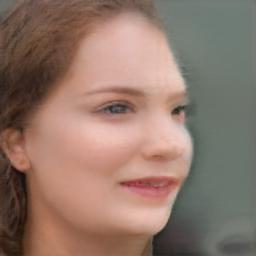}\\
    \raisebox{1.5\height}{\rotatebox[origin=c]{90}{Rendered}} &
    \includegraphics[width=\www]{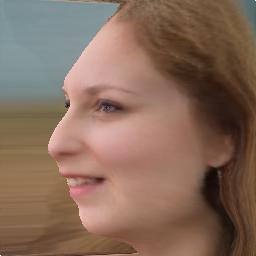}\hfill
    \includegraphics[width=\www]{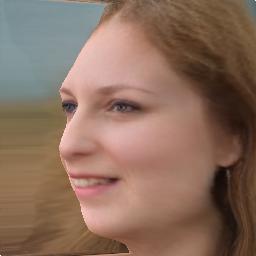}\hfill
    \includegraphics[width=\www]{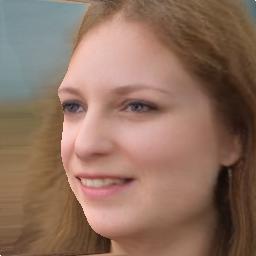}\hfill
    \includegraphics[width=\www]{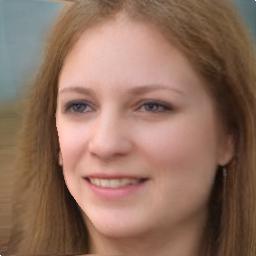}\hfill
    \includegraphics[width=\www]{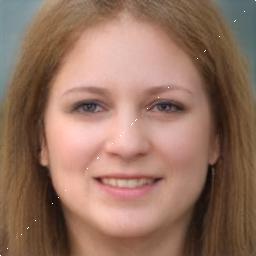}\hfill
    \includegraphics[width=\www]{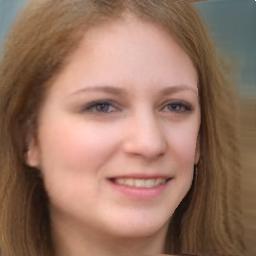}\hfill
    \includegraphics[width=\www]{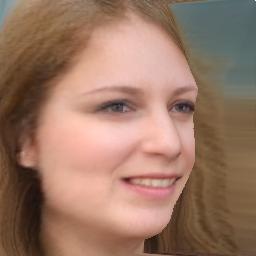}\hfill
    \includegraphics[width=\www]{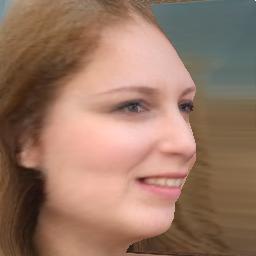}\hfill
    \includegraphics[width=\www]{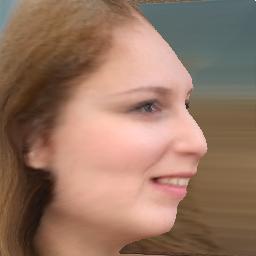}\\[0.5em]
    \raisebox{1.2\height}{\rotatebox[origin=c]{90}{StyleGAN2}} &
    \includegraphics[width=\www]{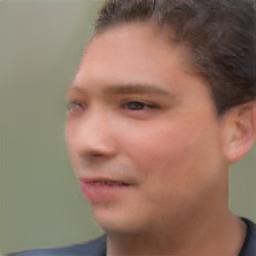}\hfill
    \includegraphics[width=\www]{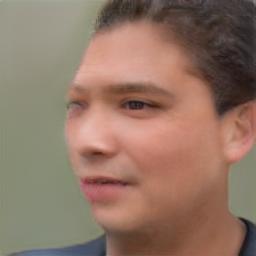}\hfill
    \includegraphics[width=\www]{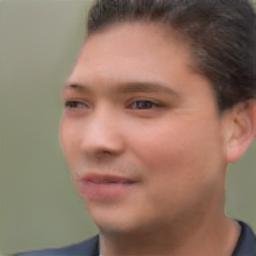}\hfill
    \includegraphics[width=\www]{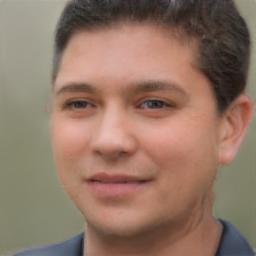}\hfill
    \includegraphics[width=\www]{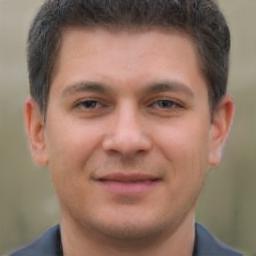}\hfill
    \includegraphics[width=\www]{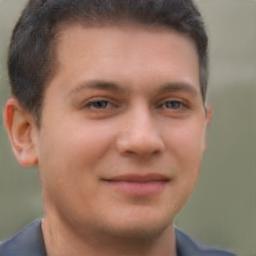}\hfill
    \includegraphics[width=\www]{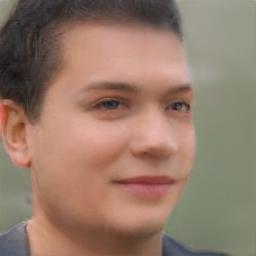}\hfill
    \includegraphics[width=\www]{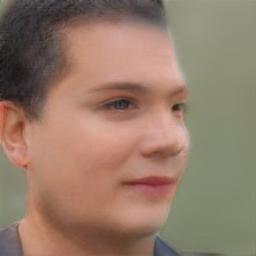}\hfill
    \includegraphics[width=\www]{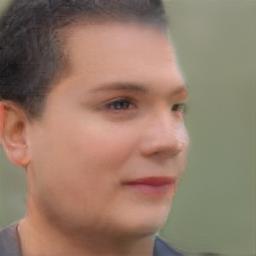}\\
    \raisebox{1.5\height}{\rotatebox[origin=c]{90}{Rendered}} &
    \includegraphics[width=\www]{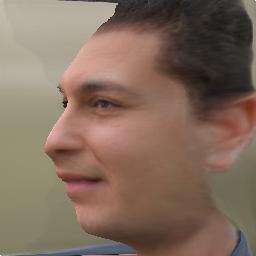}\hfill
    \includegraphics[width=\www]{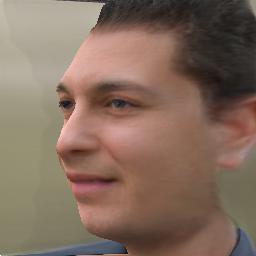}\hfill
    \includegraphics[width=\www]{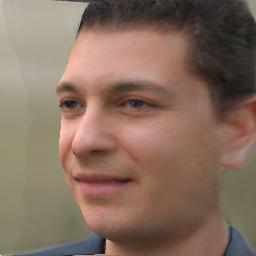}\hfill
    \includegraphics[width=\www]{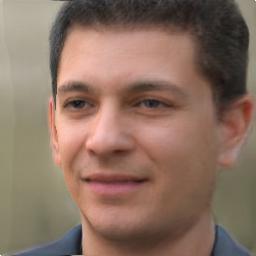}\hfill
    \includegraphics[width=\www]{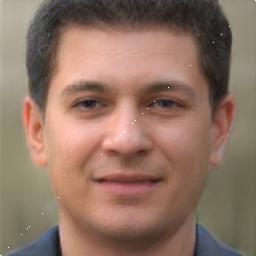}\hfill
    \includegraphics[width=\www]{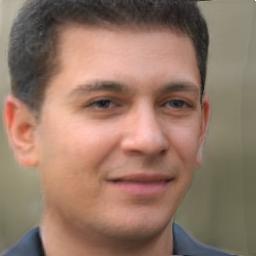}\hfill
    \includegraphics[width=\www]{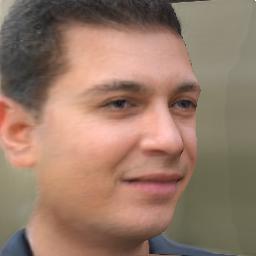}\hfill
    \includegraphics[width=\www]{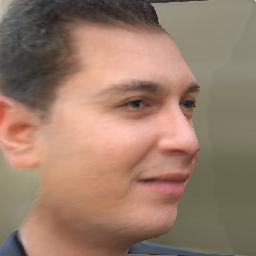}\hfill
    \includegraphics[width=\www]{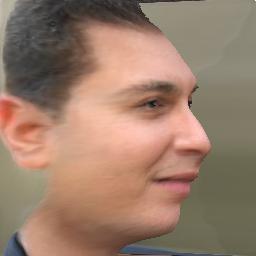}\\[0.5em]
    \raisebox{1.2\height}{\rotatebox[origin=c]{90}{StyleGAN2}} &
    \includegraphics[width=\www]{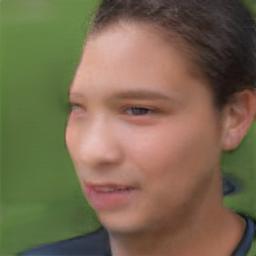}\hfill
    \includegraphics[width=\www]{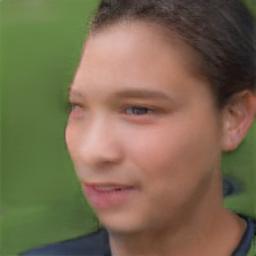}\hfill
    \includegraphics[width=\www]{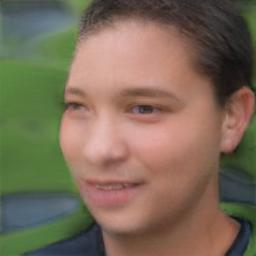}\hfill
    \includegraphics[width=\www]{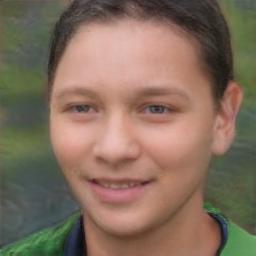}\hfill
    \includegraphics[width=\www]{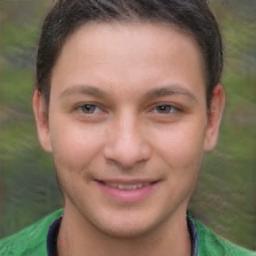}\hfill
    \includegraphics[width=\www]{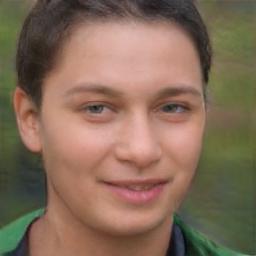}\hfill
    \includegraphics[width=\www]{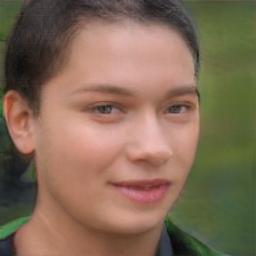}\hfill
    \includegraphics[width=\www]{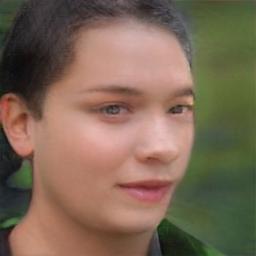}\hfill
    \includegraphics[width=\www]{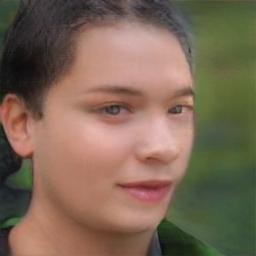}\\
    \raisebox{1.5\height}{\rotatebox[origin=c]{90}{Rendered}} &
    \includegraphics[width=\www]{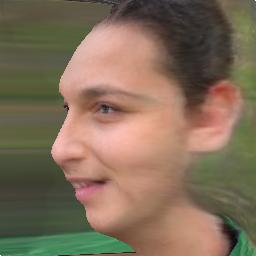}\hfill
    \includegraphics[width=\www]{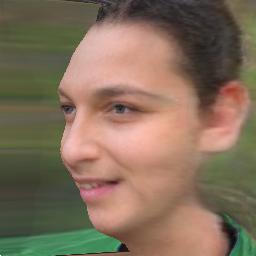}\hfill
    \includegraphics[width=\www]{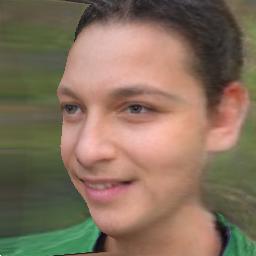}\hfill
    \includegraphics[width=\www]{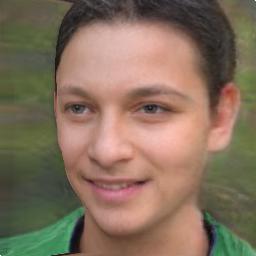}\hfill
    \includegraphics[width=\www]{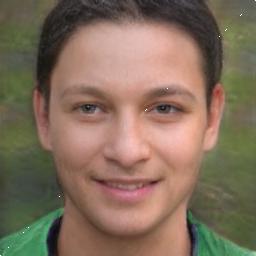}\hfill
    \includegraphics[width=\www]{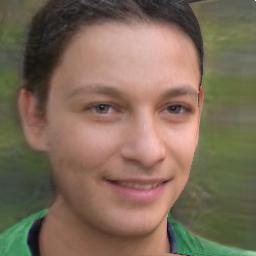}\hfill
    \includegraphics[width=\www]{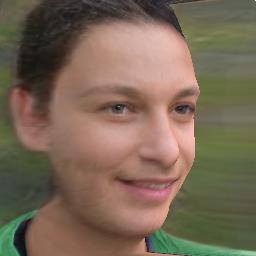}\hfill
    \includegraphics[width=\www]{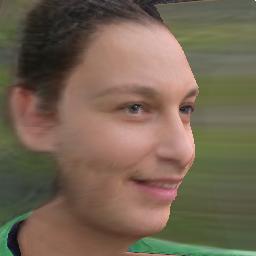}\hfill
    \includegraphics[width=\www]{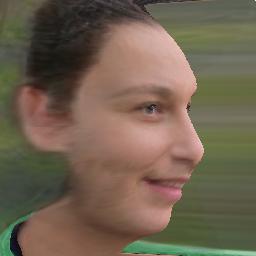}\\[0.5em]
    \raisebox{1.2\height}{\rotatebox[origin=c]{90}{StyleGAN2}} &
    \includegraphics[width=\www]{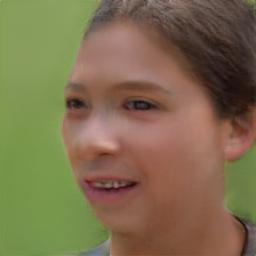}\hfill
    \includegraphics[width=\www]{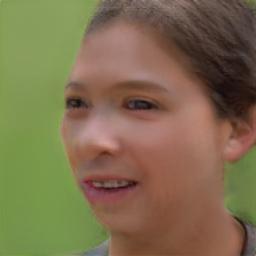}\hfill
    \includegraphics[width=\www]{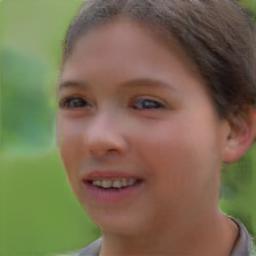}\hfill
    \includegraphics[width=\www]{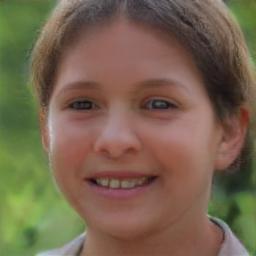}\hfill
    \includegraphics[width=\www]{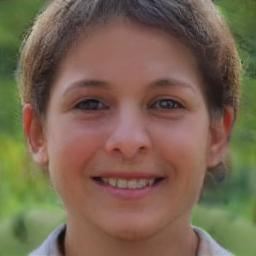}\hfill
    \includegraphics[width=\www]{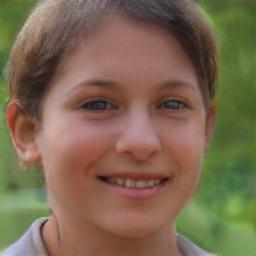}\hfill
    \includegraphics[width=\www]{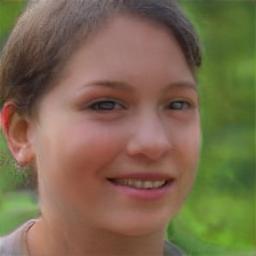}\hfill
    \includegraphics[width=\www]{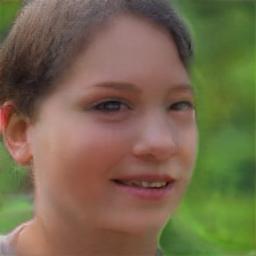}\hfill
    \includegraphics[width=\www]{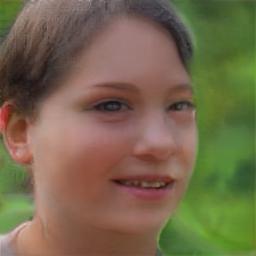}\\
    \raisebox{1.5\height}{\rotatebox[origin=c]{90}{Rendered}} &
    \includegraphics[width=\www]{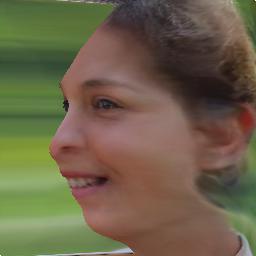}\hfill
    \includegraphics[width=\www]{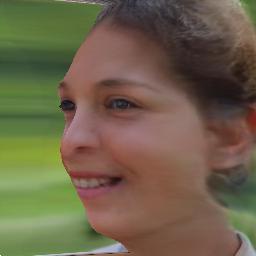}\hfill
    \includegraphics[width=\www]{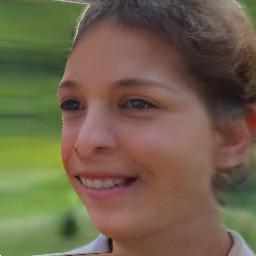}\hfill
    \includegraphics[width=\www]{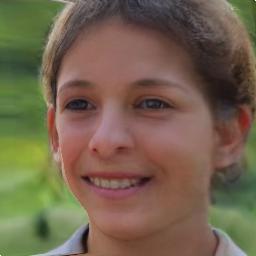}\hfill
    \includegraphics[width=\www]{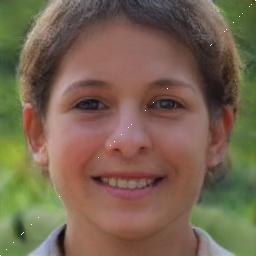}\hfill
    \includegraphics[width=\www]{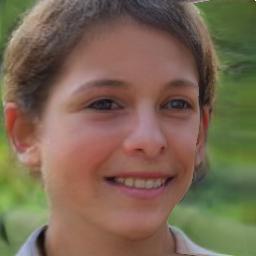}\hfill
    \includegraphics[width=\www]{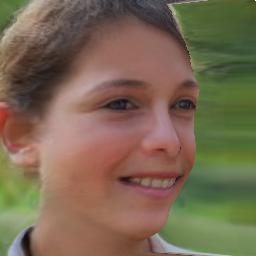}\hfill
    \includegraphics[width=\www]{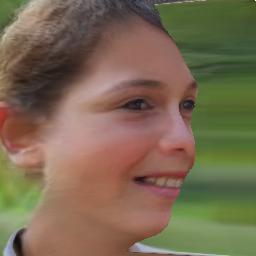}\hfill
    \includegraphics[width=\www]{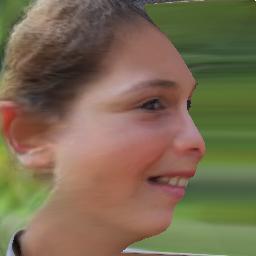}\\
\end{tabularx}
    \caption{Results of rotating faces to yaw angles of -60,-45,-30,-15,0,15,30,45,60. For each two rows, the first row shows the results of generating by manipulating the StyleGAN2 latent style code, while the second row shows the results of 3D rendered faces.}
    \label{appendix:fig:generation_multiview_yaw}
\end{figure*}

\begin{figure*}[t]
\captionsetup{font=small}
\centering
\footnotesize
\setlength\tabcolsep{1px}
\newcommand{\www}{0.108\linewidth}
\renewcommand{\arraystretch}{0.0}
\newcolumntype{Y}{>{\centering\arraybackslash}X}
\begin{tabularx}{0.8\linewidth}{Y>{\centering\arraybackslash}c}
    \raisebox{1.2\height}{\rotatebox[origin=c]{90}{StyleGAN2}} &
    \includegraphics[width=\www]{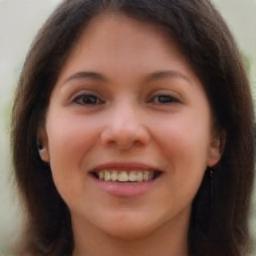}\hfill
    \includegraphics[width=\www]{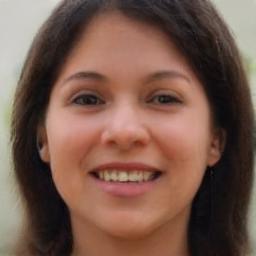}\hfill
    \includegraphics[width=\www]{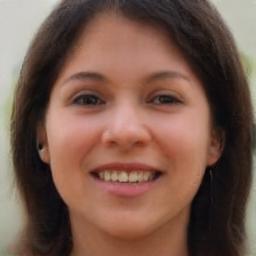}\hfill
    \includegraphics[width=\www]{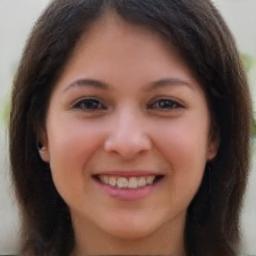}\hfill
    \includegraphics[width=\www]{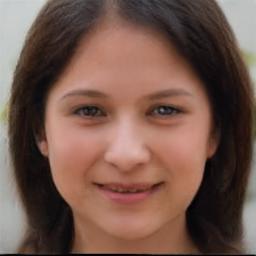}\hfill
    \includegraphics[width=\www]{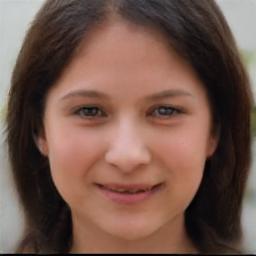}\hfill
    \includegraphics[width=\www]{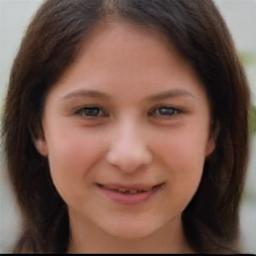}\\
    \raisebox{1.5\height}{\rotatebox[origin=c]{90}{Rendered}} &
    \includegraphics[width=\www]{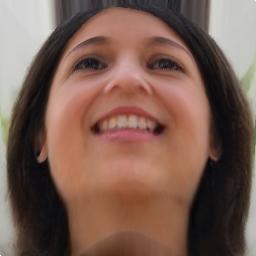}\hfill
    \includegraphics[width=\www]{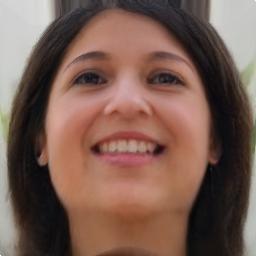}\hfill
    \includegraphics[width=\www]{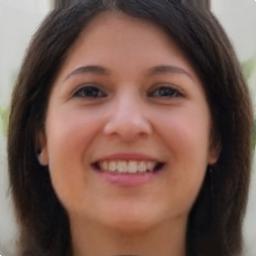}\hfill
    \includegraphics[width=\www]{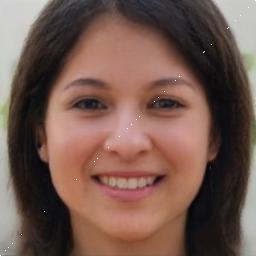}\hfill
    \includegraphics[width=\www]{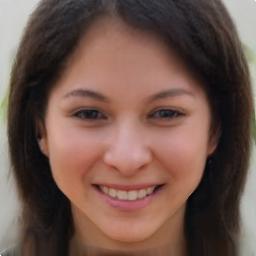}\hfill
    \includegraphics[width=\www]{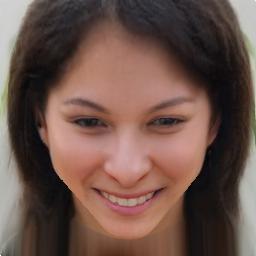}\hfill
    \includegraphics[width=\www]{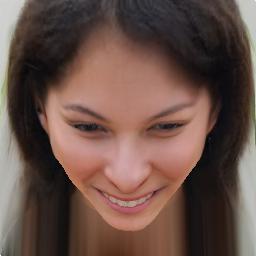}\\[0.5em]
    \raisebox{1.2\height}{\rotatebox[origin=c]{90}{StyleGAN2}} &
    \includegraphics[width=\www]{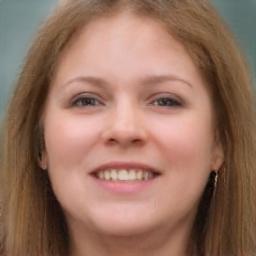}\hfill
    \includegraphics[width=\www]{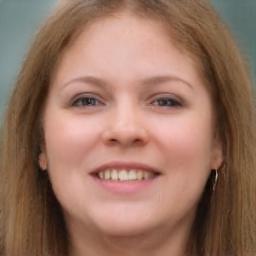}\hfill
    \includegraphics[width=\www]{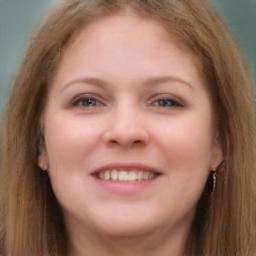}\hfill
    \includegraphics[width=\www]{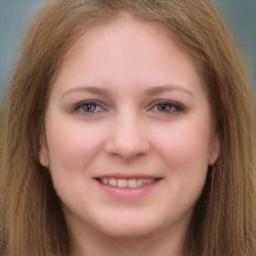}\hfill
    \includegraphics[width=\www]{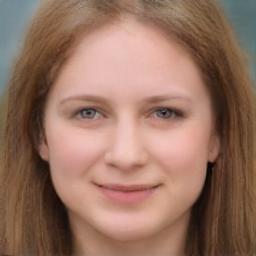}\hfill
    \includegraphics[width=\www]{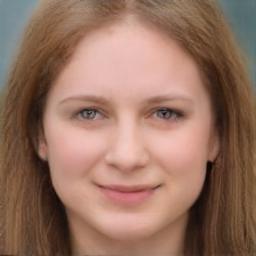}\hfill
    \includegraphics[width=\www]{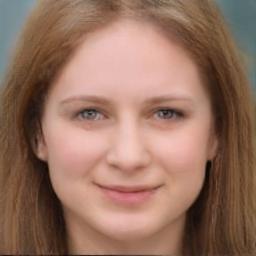}\\
    \raisebox{1.5\height}{\rotatebox[origin=c]{90}{Rendered}} &
    \includegraphics[width=\www]{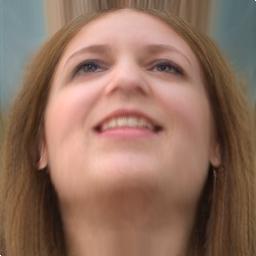}\hfill
    \includegraphics[width=\www]{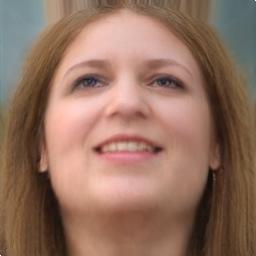}\hfill
    \includegraphics[width=\www]{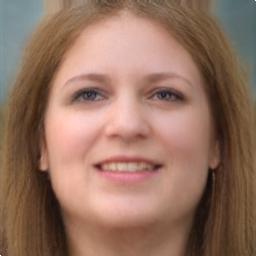}\hfill
    \includegraphics[width=\www]{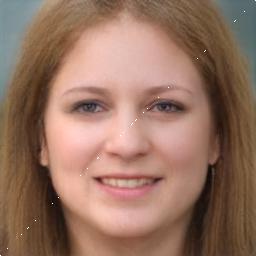}\hfill
    \includegraphics[width=\www]{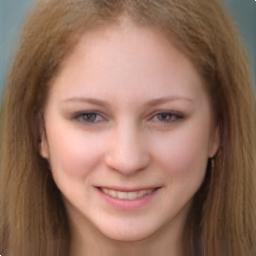}\hfill
    \includegraphics[width=\www]{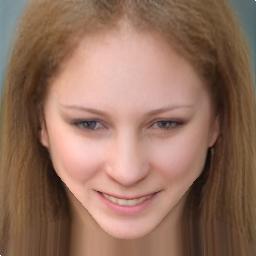}\hfill
    \includegraphics[width=\www]{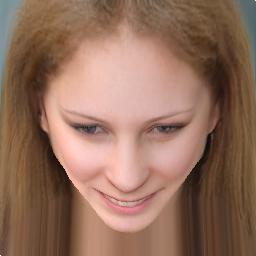}\\[0.5em]
    \raisebox{1.2\height}{\rotatebox[origin=c]{90}{StyleGAN2}} &
    \includegraphics[width=\www]{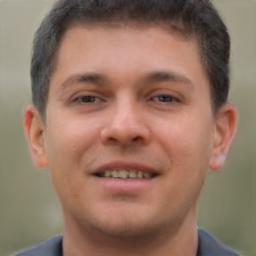}\hfill
    \includegraphics[width=\www]{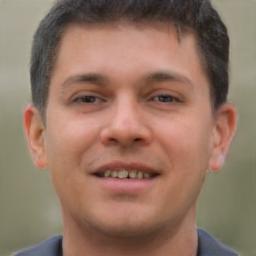}\hfill
    \includegraphics[width=\www]{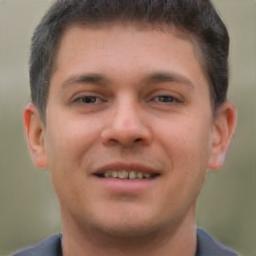}\hfill
    \includegraphics[width=\www]{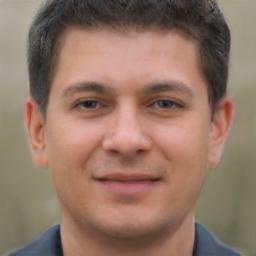}\hfill
    \includegraphics[width=\www]{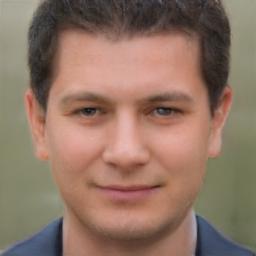}\hfill
    \includegraphics[width=\www]{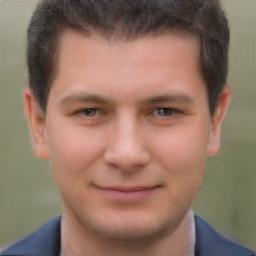}\hfill
    \includegraphics[width=\www]{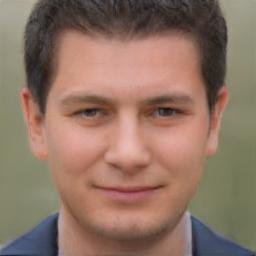}\\
    \raisebox{1.5\height}{\rotatebox[origin=c]{90}{Rendered}} &
    \includegraphics[width=\www]{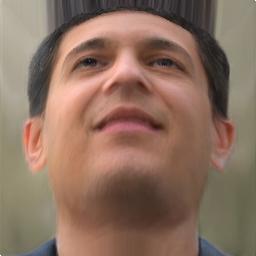}\hfill
    \includegraphics[width=\www]{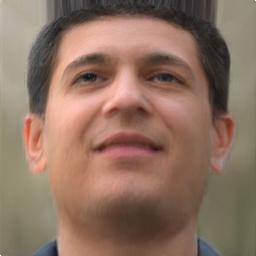}\hfill
    \includegraphics[width=\www]{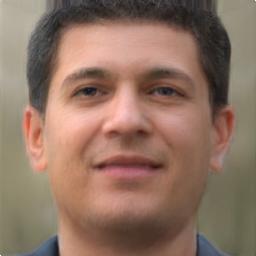}\hfill
    \includegraphics[width=\www]{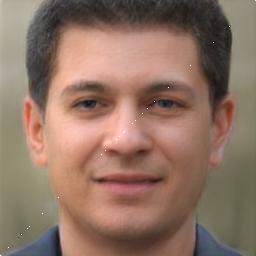}\hfill
    \includegraphics[width=\www]{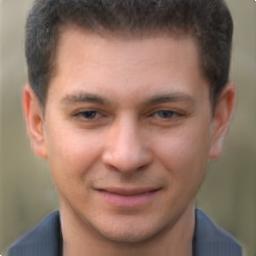}\hfill
    \includegraphics[width=\www]{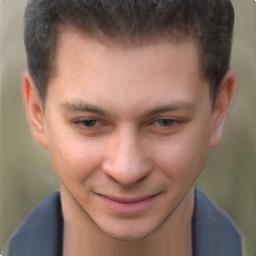}\hfill
    \includegraphics[width=\www]{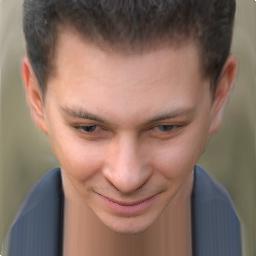}\\[0.5em]
    \raisebox{1.2\height}{\rotatebox[origin=c]{90}{StyleGAN2}} &
    \includegraphics[width=\www]{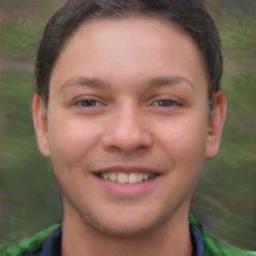}\hfill
    \includegraphics[width=\www]{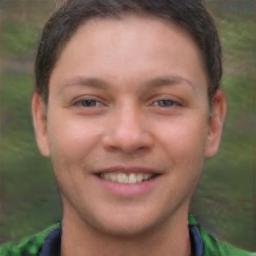}\hfill
    \includegraphics[width=\www]{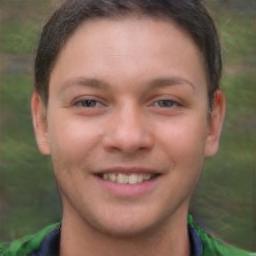}\hfill
    \includegraphics[width=\www]{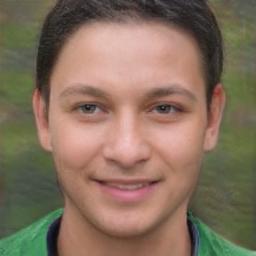}\hfill
    \includegraphics[width=\www]{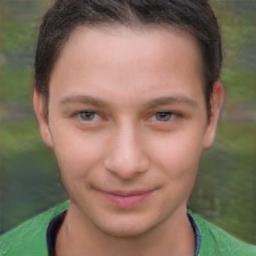}\hfill
    \includegraphics[width=\www]{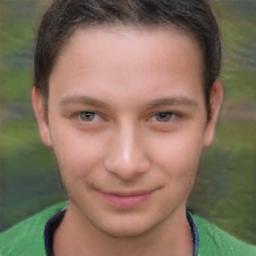}\hfill
    \includegraphics[width=\www]{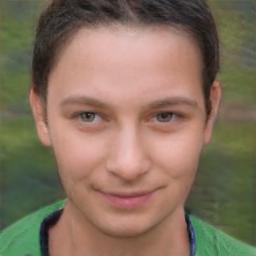}\\
    \raisebox{1.5\height}{\rotatebox[origin=c]{90}{Rendered}} &
    \includegraphics[width=\www]{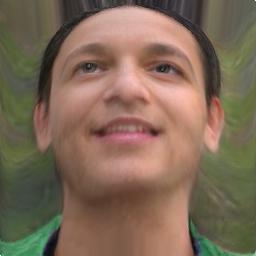}\hfill
    \includegraphics[width=\www]{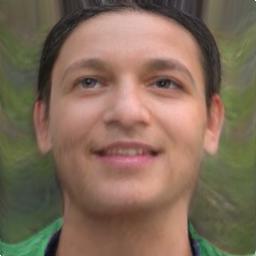}\hfill
    \includegraphics[width=\www]{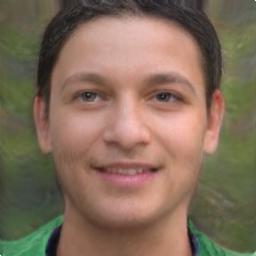}\hfill
    \includegraphics[width=\www]{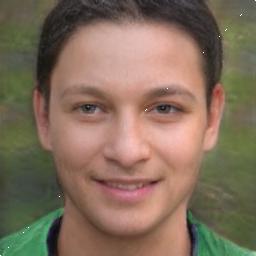}\hfill
    \includegraphics[width=\www]{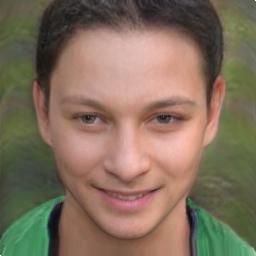}\hfill
    \includegraphics[width=\www]{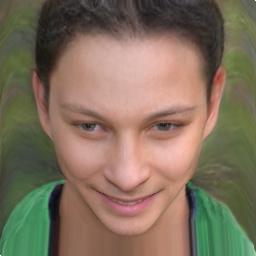}\hfill
    \includegraphics[width=\www]{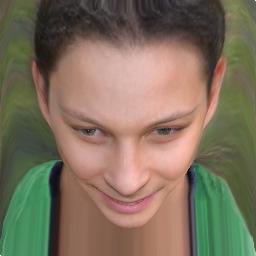}\\[0.5em]
    \raisebox{1.2\height}{\rotatebox[origin=c]{90}{StyleGAN2}} &
    \includegraphics[width=\www]{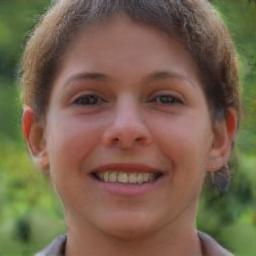}\hfill
    \includegraphics[width=\www]{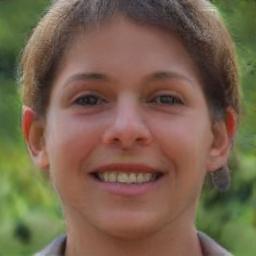}\hfill
    \includegraphics[width=\www]{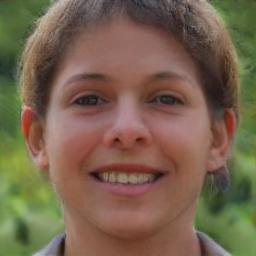}\hfill
    \includegraphics[width=\www]{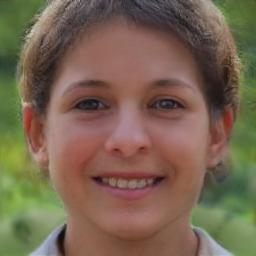}\hfill
    \includegraphics[width=\www]{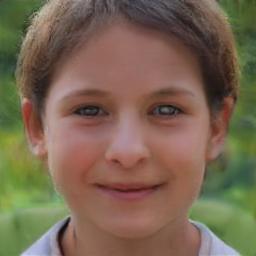}\hfill
    \includegraphics[width=\www]{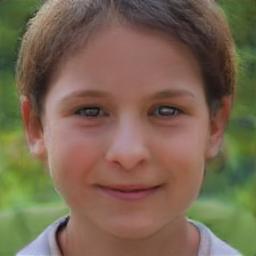}\hfill
    \includegraphics[width=\www]{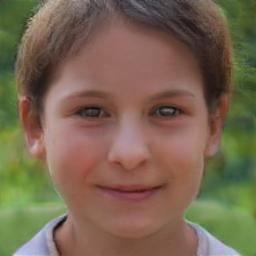}\\
    \raisebox{1.5\height}{\rotatebox[origin=c]{90}{Rendered}} &
    \includegraphics[width=\www]{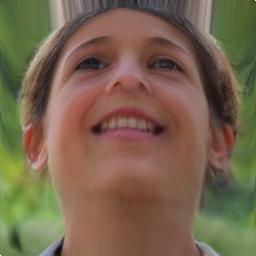}\hfill
    \includegraphics[width=\www]{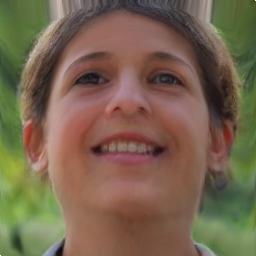}\hfill
    \includegraphics[width=\www]{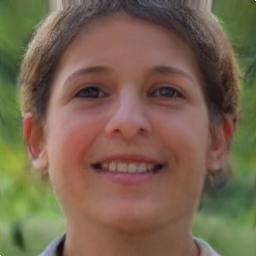}\hfill
    \includegraphics[width=\www]{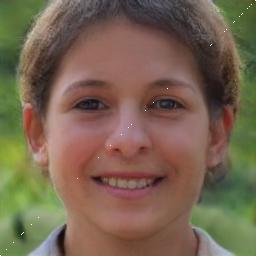}\hfill
    \includegraphics[width=\www]{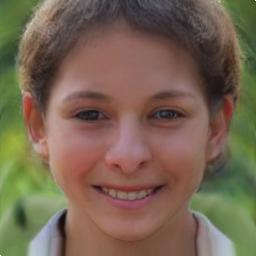}\hfill
    \includegraphics[width=\www]{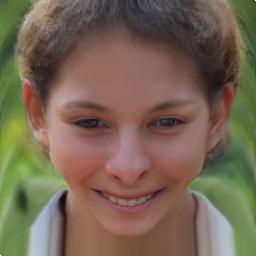}\hfill
    \includegraphics[width=\www]{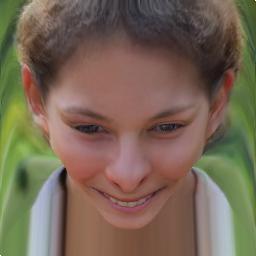}\\[0.5em]
\end{tabularx}
    \caption{Results of rotating faces to pitch angles of -30,-20,-10,0,10,20,30. For each two rows, the first row shows the results of generating by manipulating the StyleGAN2 latent style code, while the second row shows the results of 3D rendered faces.}
    \label{appendix:fig:generation_multiview_pitch}
\end{figure*}

\end{document}